\documentclass[sigconf]{acmart}

\AtBeginDocument{%
  \providecommand\BibTeX{{%
    \normalfont B\kern-0.5em{\scshape i\kern-0.25em b}\kern-0.8em\TeX}}}

\setcopyright{iw3c2w3}
\copyrightyear{2021}
\acmYear{2021}
\acmDOI{10.1145/3442381.3449897}

\acmConference[WWW '21]{Proceedings of the Web Conference 2021}{April 19--23, 2021}{Ljubljana, Slovenia}
\acmBooktitle{Proceedings of the Web Conference 2021 (WWW '21), April 19--23, 2021,
Ljubljana, Slovenia}
\acmPrice{}
\acmISBN{978-1-4503-8312-7/21/04}
\usepackage{booktabs}
\usepackage{multirow}
\usepackage{pifont}
\usepackage{bm}

\begin{document}

\title{Boosting the Speed of Entity Alignment \textbf{$10\times$}: Dual Attention Matching Network with Normalized Hard Sample Mining}

\author{Xin Mao$^{1*}$, Wenting Wang$^2$, Yuanbin Wu$^1$, Man Lan$^{1*}$}
\email{xmao@stu.ecnu.edu.cn, {wenting.wang}@lazada.com, {ybwu,mlan}@cs.ecnu.edu.cn}
\affiliation{$^1$East China Normal University, $^2$Alibaba Group}

\begin{abstract}
Seeking the equivalent entities among multi-source Knowledge Graphs (KGs) is the pivotal step to KGs integration, also known as \emph{entity alignment} (EA).
However, most existing EA methods are inefficient and poor in scalability.
A recent summary points out that some of them even require several days to deal with a dataset containing $200,000$ nodes (DWY$100$K).
We believe over-complex graph encoder and inefficient negative sampling strategy are the two main reasons.
In this paper, we propose a novel KG encoder --- \emph{Dual Attention Matching Network} (Dual-AMN), which not only models both intra-graph and cross-graph information smartly, but also greatly reduces computational complexity.
Furthermore, we propose the \emph{Normalized Hard Sample Mining Loss} to smoothly select hard negative samples with reduced loss shift.
The experimental results on widely used public datasets indicate that our method achieves both high accuracy and high efficiency.
On DWY$100$K, the whole running process of our method could be finished in $1,100$ seconds, at least $10\times$ faster than previous work.
The performances of our method also outperform previous works across all datasets, where $Hits@1$ and $MRR$ have been improved from $6\%$ to $13\%$.
\end{abstract}



\maketitle

\section{Introduction}
\label{sec:intro}
Typically, knowledge graphs (KGs) store real-world knowledge in the form of triples (i.e., $<$entity, relation, entity$>$), where entities are connected through relations.
In recent years, many general KGs (e.g., DBpedia \cite{DBLP:conf/semweb/AuerBKLCI07}, YAGO \cite{DBLP:conf/www/SuchanekKW07}) and domain-specific KGs (e.g., Scientific \cite{DBLP:conf/kdd/TangZYLZS08}) have proliferated
and been widely used in downstream applications, such as search engines and recommendation systems.

\begin{figure}
  \centering
  \includegraphics[width=1\linewidth]{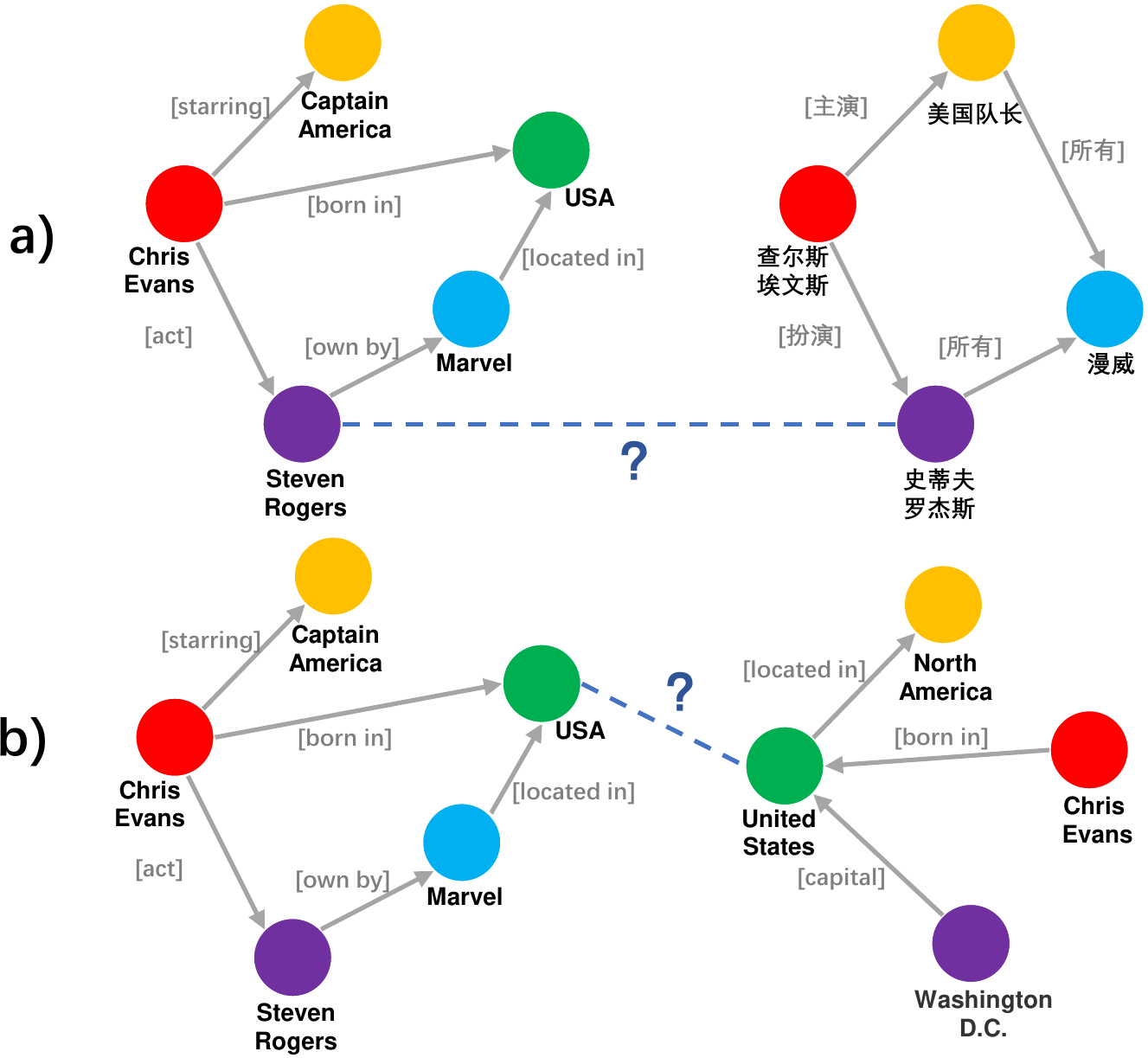}
  \caption{Two examples of KGs integration. Figure (a) represents cross-lingual and Figure (b) represents cross-domain.}\label{Figure:Intro}
\end{figure}

In practice, a KG is usually constructed from one single data source.
Therefore, it is unlikely to cover the full domain.
As shown in Figure \ref{Figure:Intro}(a), integrating KGs in the same domain but built from different languages can transfer the information from high resource language to low resource language.
This in turn will facilitate downstream cross-lingual applications, especially for minority language users.
Moreover, consolidating multi-domain KGs (Figure \ref{Figure:Intro}(b)) can supplement cross-domain information and improve the coverage, thus making KGs more complete.

Seeking the equivalent entities among multi-source KGs is the pivotal step to KGs integration, also known as \emph{entity alignment} (EA).
Recently, EA attracts enormous attention and progresses rapidly.
Dozens of related papers have been published in recent years.
In general, these methods all share one core framework:
assume that equivalent entities possess similar neighboring structure, apply \emph{KG embedding} methods (e.g., TransE \cite{DBLP:conf/nips/BordesUGWY13} or GCN \cite{DBLP:journals/corr/KipfW16}) to obtain dense embeddings for each entity, then map these embeddings into a unified vector space by \emph{alignment module} (e.g., Triplet loss and Contrastive Loss\cite{DBLP:conf/cvpr/SchroffKP15,DBLP:conf/cvpr/HadsellCL06}),
and finally the pair-wise distance between entities determines whether they are aligned or not.

However, previous EA methods are inefficient and poor in scalability, as summarized by \citet{9174835} that most of them require several hours \cite{DBLP:conf/emnlp/LiCHSLC19,DBLP:conf/ijcai/SunHZQ18,DBLP:conf/acl/CaoLLLLC19} or even days \cite{DBLP:conf/acl/XuWYFSWY19} on a dataset containing $200,000$ nodes (i.e., DWY$100$K).
In reality, KGs usually consist of millions of entities and relations (e.g., the full DBpedia contains $10+$ billion entities, $1+$ trillion triples).
Such large-scale datasets impose huge challenges in the efficiency and scalability of EA methods.
Obviously, the high cost in time hinders the feasibility of applying these EA methods to large-scale KGs.

We believe there are two main reasons that cause the high time complexity of these advanced methods:

(1) \textbf{Over-complex graph encoder}:
Since vanilla GCN is unable to model the heterogeneous relation information in KGs, many relation-aware GNN variants are proposed in EA task.
However, some GNN variants are over-complex and inefficient.
The running time of the vanilla GCN \cite{DBLP:conf/emnlp/WangLLZ18} is only $10\%$ of those from complex encoders.
Every time a complex technique is introduced, e.g., Graph Attention mechanism \cite{DBLP:conf/iclr/VelickovicCCRLB18}, Graph Matching Networks \cite{DBLP:conf/icml/LiGDVK19} (GMN), Joint Learning \cite{DBLP:conf/emnlp/LiCHSLC19}, the time complexity is dramatically increased.
For instance, GM-Align \cite{DBLP:conf/acl/XuWYFSWY19} incorporates GMN and achieves decent performances on a small dataset (DBP$15$K), but the performance improvement most likely is contributed from the literal information.
When moving to a larger dataset (DWY$100$K), GM-Align needs five days to obtain the results.
We believe the graph encoder still has some redundancy in design and its architecture can be further simplified to reduce time consumption.

(2) \textbf{Inefficient negative sampling strategy}:
Almost all existing EA methods rely on the pair-wise loss functions (e.g., TransE, Triplet loss, and Contrastive Loss).
In pair-wise loss, the negative samples are constructed via uniform random sampling.
In this way, the samples are usually highly redundant and have limited information.
The learning process could be hampered by the low-quality negative samples, resulting in slow convergence and model degradation.
To alleviate this problem, BootEA \cite{DBLP:conf/ijcai/SunHZQ18} proposes a \emph{Truncated Uniform Negative Sampling} strategy to choose K-nearest neighbors as negative samples (i.e., hard samples).
Such an intuitive and effective strategy has been widely adopted in subsequent studies \cite{DBLP:conf/emnlp/LiCHSLC19,DBLP:journals/corr/abs-2004-13579,DBLP:conf/acl/CaoLLLLC19}.
However, ranking all neighbors to find the K-nearest is highly time-consuming and difficult to be fully parallelized on GPU.
For example, \emph{Truncated Uniform Negative Sampling Strategy} takes more than $25\%$ of the whole time cost of BootEA.

Instead of trading efficiency for better performance, in this paper, we propose \emph{Dual Attention Matching Network} (Dual-AMN) to capture dual relational information within a single graph and across two graphs:
The \emph{Simplified Relational Attention Layer} captures relational information within each KG by generating relation-specific embeddings through \emph{Relational Anisotropy Attention} and \emph{Relational Projection}.
The \emph{Proxy Matching Attention Layer} treats alignment as a special relation type and explicitly models it via proxy vectors.
In addition, to tackle the inefficient sampling issue, we further propose a \emph{Normalized Hard Sample Mining Loss}.
First, \emph{LogSumExp} operation is used to approximate \emph{Max} operation to generate hard samples smoothly but efficiently.
Then, to resolve the dilemma of hyper-parameter selection in \emph{LogSumExp}, we introduce a loss normalization strategy adjusting the distribution of loss dynamically.

Experiment with the same hardware environment, our method could finish the whole running process in $1,100$ seconds on DWY-$100$K, including data loading, training, and evaluating, which is $3\times$ faster compared to the fastest existing model (i.e., GCN-Align \cite{DBLP:conf/emnlp/WangLLZ18}) and only takes up $10\%$ of advanced methods.
On DBP$15$K with a smaller scale, our method even could obtain results in less than $40$ seconds.
More surprisingly, the alignment results obtained by our method have very high accuracy.
The experiments show that our method beats all state-of-the-art competitors across all datasets, and the performance improvement on $hits@1$ and $MRR$ ranges from $6\%$ to $13\%$.
The main contributions are summarized as follows:
\begin{itemize}
  \item \textbf{Model.}
  We propose a novel graph encoder  \emph{Dual Attention Matching Network} (Dual-AMN) composing of \emph{Simplified Relational Attention Layer} and \emph{Proxy Matching Attention Layer}.
  The proposed encoder not only models both intra-graph and cross-graph relations smartly, but also greatly reduces computational complexity.
  \item \textbf{Training.}
  Instead of the inefficient sampling strategy, we propose a \emph{Normalized Hard Sample Mining Loss}, where the \emph{LogSumExp} operation generates hard samples efficiently and the loss normalization alleviates the dilemma of hyper-parameter selection.
  The new loss dramatically cuts down the sampling consumption and accelerates the convergence speed of the model.
  \item \textbf{Experiments.}
  The experimental results on widely used public datasets indicate that our method has high efficiency and accuracy.
  Furthermore, we design many auxiliary experiments to demonstrate the effectiveness of each component and the interpretability of the model.
\end{itemize}

\section{Task Definition}
\textbf{\emph{Definition of Knowledge Graph:}}
The formal definition of a KG is a directed graph $G=(E,R,T)$ comprising three sets --- entities $E$, relations $R$, and triples $T\subseteq E\times R\times E$.
KG stores the real-world information in the form of triples $<$entity, relation, entity$>$, which describe the inherent relation between two entities.
In addition, we define $\mathcal{N}_{e_i}$ to represent the neighbor set of entity $e_i$ and $\mathcal{R}_{ij}$ represent the set of relations between $e_i$ and $e_j$.

\noindent
\textbf{\emph{Definition of Entity Alignment:}}
Given two KGs $G_1 = (E_1,R_1,T_1)$, $G_2 = (E_2,R_2,T_2)$, and a pre-aligned entity pair set $P = \{(u,v)|u\in E_1, u\in E_2\, u\equiv v\}$, where $\equiv$ denotes equivalence.
EA aims to obtain more potential equivalent entity pairs based on the information of $G_1$, $G_2$, and $P$.

\section{Related Work}
\label{sec:rw}
As mentioned in Section \ref{sec:intro}, existing EA methods can be abstracted into one framework containing three major components:
\begin{itemize}
  \item \textbf{Graph embedding module} is responsible for encoding entities and relations of KGs into dense embeddings.
  \item \textbf{Entity alignment module} aims to map the embeddings of multi-source KGs into a unified vector space via pre-aligned entity pairs.
  \item \textbf{Information enhancement module} is able to generate semi-supervised data or introduce additional literal information for enhancement.
\end{itemize}
In this section, we categorize existing EA approaches based on their designs of these three components, as shown in Table \ref{tabel:rw}.

\begin{table}
\begin{center}

\resizebox{1\linewidth}{!}{
\renewcommand\arraystretch{1.6}
\large
\begin{tabular}{cccc}
  \toprule
  \textbf{Method}&\textbf{Embedding}&\textbf{Alignment}&\textbf{Enhancement}\\
  \toprule
  MTransE \cite{DBLP:conf/ijcai/ChenTYZ17} &TransE&Mapping&None\\
  GCN-Align \cite{DBLP:conf/emnlp/WangLLZ18}&GNN&Margin-based&None\\
  RSNs \cite{DBLP:conf/icml/GuoSH19}&RSNs&Corpus fusion&None\\
  MuGNN \cite{DBLP:conf/acl/CaoLLLLC19}&Hybrid&Margin-based&None\\
  KECG \cite{DBLP:conf/emnlp/LiCHSLC19}&Hybrid&Margin-based&None\\
  \midrule
  BootEA \cite{DBLP:conf/ijcai/SunHZQ18}&TransE&Corpus fusion&Semi-supervised\\
  NAEA \cite{DBLP:conf/ijcai/ZhuZ0TG19}&Hybrid&Corpus fusion&Semi-supervised\\
  TransEdge\cite{DBLP:journals/corr/abs-2004-13579}&TransE&Corpus fusion&Semi-supervised\\
  MRAEA \cite{DBLP:conf/wsdm/MaoWXLW20}&GNN&Margin-based&Semi-supervised\\
  \midrule
  GM-Align \cite{DBLP:conf/acl/XuWYFSWY19}&GNN&Margin-based&Entity Name\\
  RDGCN \cite{DBLP:conf/ijcai/WuLF0Y019}&GNN&Margin-based&Entity Name\\
  HMAN \cite{DBLP:conf/emnlp/YangZSLLS19}&GNN& Margin-based&Attribute\\
  HGCN \cite{DBLP:conf/emnlp/WuLFWZ19}&GNN&Margin-based&Entity Name\\
  \bottomrule
\end{tabular}
}
\end{center}
\caption{Categorization of some popular EA approaches.}\label{tabel:rw}
\end{table}

\subsection{Embedding Module}
TransE \cite{DBLP:conf/nips/BordesUGWY13}, GNN, and Hybrid are the three mainstream embedding approaches.
TransE interprets relations as the translation from head entities to tail entities and assumes that the embeddings of entities and relations follow the assumption $\bm{h} + \bm{r} \approx \bm{t}$ if a triple $(h,r,t)$ holds.
Based on this hypothesis, many variants (e.g., TransH \cite{DBLP:conf/aaai/WangZFC14} and TransR \cite{DBLP:conf/aaai/LinLSLZ15}) are proposed and proven to be effective in subsequent studies.
Graph Neural Network (GNN) is famous for its strong modeling capability on the non-Euclidean structure.
Different from TransE optimizing triples, GNN generates node-aware embeddings by aggregating the neighboring information of entities.
However, vanilla GNN \cite{DBLP:journals/corr/KipfW16} is unable to encode heterogeneous relational graphs such as KGs.
Thus, many subsequent studies focus on modifying GNN to fit into KG.
The main direction is to use the anisotropic attention mechanism \cite{DBLP:conf/iclr/VelickovicCCRLB18} to assign different weight coefficients to entities.
A GNN model whose node update equation treats every edge direction equally, is considered isotropic (e.g., vanilla GCN); and a GNN model whose node update equation treats every edge direction differently, is considered anisotropic (e.g., GAT \cite{DBLP:conf/iclr/VelickovicCCRLB18}).
Hybrid embedding approaches combine TransE and GNN together, which aim to enhance the expression ability of the model.
However, for now, the best-performing methods TransEdge \cite{DBLP:journals/corr/abs-2004-13579} and MRAEA \cite{DBLP:conf/wsdm/MaoWXLW20} are not hybrid.
The hybrid-based methods do not show necessity while introducing additional complexity.

In addition to these three mainstream approaches, RSNs \cite{DBLP:conf/icml/GuoSH19} integrates Recurrent Neural Networks (RNNs) with a skipping mechanism to efficiently capture the long-term relational dependencies within and between KGs.
RSNs performs well on sparse KGs, but it is still weaker than SOTA mainstream methods.

\subsection{Alignment Module}
The most common alignment methods are as follows:
(1) \textbf{Mapping} \cite{DBLP:conf/ijcai/ChenTYZ17} uses one or two linear transformation matrices to map the embeddings of entities in different KGs into a unified vector space.
This idea is inspired by the cross-lingual word embedding task \cite{DBLP:conf/iclr/LampleCRDJ18}, and the first proposed EA method \cite{DBLP:conf/ijcai/ChenTYZ17} adopts this alignment module.
(2) \textbf{Corpus fusion} \cite{DBLP:conf/ijcai/SunHZQ18} swaps the entities in the pre-aligned set and generates new triples to calibrate the embeddings into a unified space.
For example, there are two triples $(e_1,r_1,e_2)\in KG_1$ and $(e_3,r_2,e_4)\in KG_2$.
If $e_1 \equiv e_3$ holds, \emph{Corpus fusion} adds two extra triples $(e_3,r_1,e_2)$ and $(e_1,r_2,e_4)$.
This approach not only integrates two KGs into one KG but also plays the role of data augmentation.
(3) \textbf{Margin-based} represents a series of pair-wise margin-based loss functions, such as Triplet loss \cite{DBLP:conf/cvpr/SchroffKP15}, Contrastive loss \cite{DBLP:conf/cvpr/HadsellCL06}, and so on.
Margin-based loss functions are often combined with Siamese Neural Network in ranking tasks (e.g., face recognition and text similarity).
Actually, GNN-based EA methods are inspired by the Siamese Neural Network and have similar architecture, so most of them use Margin-based loss to be their alignment module.

\subsection{Enhancement Module}
Because manually aligning entities is expensive in practice, pre-aligned pairs are usually a small part of all entities.
Therefore, existing methods usually reserve $30\%$ or even less of the aligned pairs as training data to simulate this situation.
Due to the lack of labeled data, some EA methods \cite{DBLP:conf/ijcai/SunHZQ18, DBLP:journals/corr/abs-2004-13579} adopt bootstrapping to generate semi-supervised data iteratively.
Based on the asymmetric nature of cross-KG alignment, MRAEA \cite{DBLP:conf/wsdm/MaoWXLW20} further proposes a bi-directional iterative strategy.
These data augmentation techniques have been proved effective in improving alignment performance.

In addition to structure, some methods \cite{DBLP:conf/emnlp/WuLFWZ19,DBLP:conf/emnlp/YangZSLLS19} propose that introducing literal information could provide a multi-aspect view for alignment models and improve accuracy.
However, it should be noted that not all datasets contain literal information, especially in practical applications.
For example, there are privacy risks when using User Generated Content (UGC).
Compared with the literal methods, the structure-only methods are more general.
Therefore, these literal methods should be compared among themselves.

\begin{figure*}[t]
  \centering
  \includegraphics[width=0.85\textwidth]{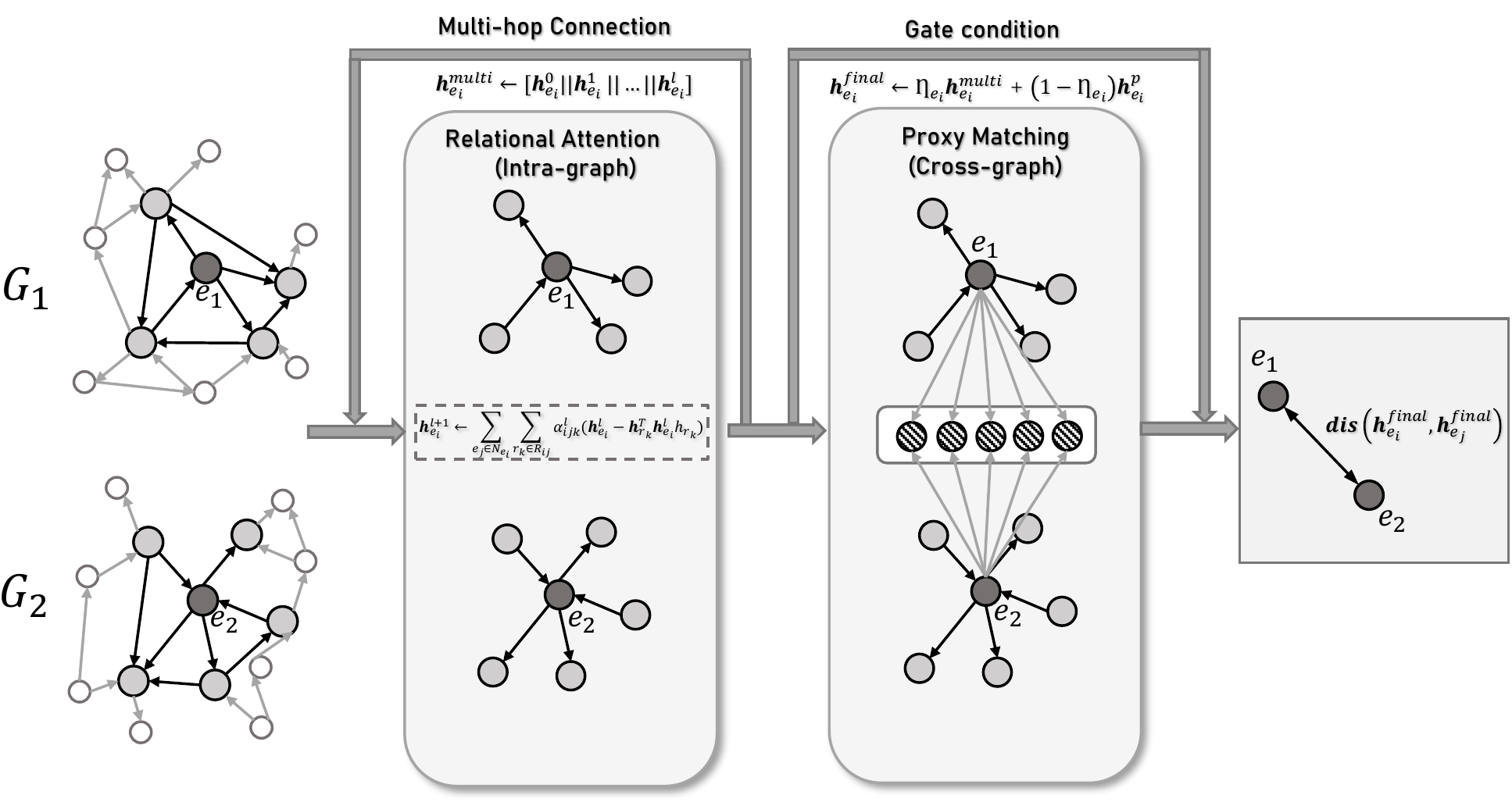}
  \caption{The architecture illustration of \emph{Dual Attention Matching Network} (Dual-AMN), composing of \emph{Simplified Relational Attention Layer} and \emph{Proxy Matching Attention Layer}.}\label{Figure:model}
\end{figure*}

\section{Dual Attention Matching Network}
\label{sec:model}
As mentioned in Section \ref{sec:intro}, existing graph encoders, which are over-complex in certain designs and poor in scalability, are not suitable to be applied to large-scale KG.
To address these defects, we propose \emph{Dual Attention Matching Network} (Dual-AMN).
Figure \ref{Figure:model} depicts that Dual-AMN is composed of two major components: \emph{Simplified Relational Attention Layer} and \emph{Proxy Matching Attention Layer}.
The \emph{Simplified Relational Attention Layer} captures relational information within each KG by generating relation-specific embeddings through \emph{Relational Anisotropy Attention} and \emph{Relational Projection}.
By treating alignment as a special relation, our \emph{Proxy Matching Attention Layer} leverages a list of proxies to explicitly capture the cross-graph information.
By combining the outcomes of these two proposed components, our Dual-AMN not only embeds both intra-graph and cross-graph relations smartly, but also greatly reduces computational complexity.
The experimental results show that the proposed method achieves the SOTA in both performance and efficiency.
In this section, we describe the architecture of Dual-AMN in details.

\subsection{Simplified Relational Attention Layer}
Since vanilla GCN is unable to model the heterogeneous relation information in KGs, many relation-aware GNN variants are proposed in EA task.
Most of them could be described by the following equation:
\begin{equation}
  \bm{h}^{l+1}_{e_i} = \sum_{e_j\in \mathcal{N}_{e_i} \cup \{e_i\}} \alpha_{ij} \bm{W}\bm{h}^{l}_{e_j}
\end{equation}
where $\bm{h}^{l}_{e_i}$ represents the embedding vector of $e_i$ obtained by the $l$-th GNN layer, $\alpha_{i,j}$ represents the weight coefficient between $e_i$ and $e_j$, $\bm{W}$ represents the transformation matrix.
Table \ref{tabel:ra} lists some popular GNN encoders.
We summarize three findings:
(1) Except GCN-Align which first utilizes GCN in EA, all the other methods adopt anisotropic attention mechanism.
This indicates that it is necessary to distinguish the importance of entities.
(2) There is a tendency that more recent methods are not joint learning based, probably because joint methods are not superior in performance.
For example, MRAEA and TransEdge outperform MuGNN, KECG, and NAEA.
So joint learning which introduces extra computation complexity is not necessary.
(3) We also notice that many methods constrain the transformation matrix $\bm{W}$ of GNN layer to be diagonal or even remove $\bm{W}$ in order to avoid performance degradation.
We believe the main reason is that the entity embeddings are all trainable and the standard linear transformation may introduce too many parameters, causing over-fitting when updating these embeddings.
Inspired by these findings, we design a simplified relation-aware GNN layer.

\begin{table}
\begin{center}
\resizebox{0.9\linewidth}{!}{
\renewcommand\arraystretch{1.1}
\begin{tabular}{cccc}
  \toprule
  \textbf{Method}&$\alpha_{ij}$&Joint&$\bm{W}$\\
  \toprule
  GCN-Align \cite{DBLP:conf/emnlp/WangLLZ18}&Isotropic&$\times$&None\\
  MuGNN \cite{DBLP:conf/acl/CaoLLLLC19}&Anisotropy &$\checkmark$&Diagonal\\
  KECG \cite{DBLP:conf/emnlp/LiCHSLC19}&Anisotropy &$\checkmark$&Diagonal\\
  NAEA \cite{DBLP:conf/ijcai/ZhuZ0TG19}&Anisotropy &$\checkmark$&Normal\\
  HMAN \cite{DBLP:conf/emnlp/YangZSLLS19}&Anisotropy &$\times$&Diagonal\\
  MRAEA \cite{DBLP:conf/wsdm/MaoWXLW20}&Anisotropy &$\times$&None\\
  \bottomrule
\end{tabular}
}
\end{center}
\caption{Categorization of GNN encoders in some popular EA approaches. }\label{tabel:ra}
\end{table}

The inputs of our model are two metrics, $\bm{H}^e \in \mathbb{R}^{|E|\times d}$ represents the initial entity features
and $\bm{H}^r \in \mathbb{R}^{|R|\times d}$ represents the initial relation features.
Both of them are randomly initialized by \emph{He\_initializer} \cite{DBLP:conf/iccv/HeZRS15}.
Similar to existing EA methods, we use anisotropic relational attention mechanism to aggregate the neighborhood information around entities.
The output embedding of entity $e_i$ at the $l$-th layer is obtained by the following equation:
\begin{equation}
  \bm{h}^{l+1}_{e_i} = tanh\left(\sum_{e_j\in \mathcal{N}_{e_i}} \sum_{r_k \in \mathcal{R}_{ij}} \alpha_{ijk}^l (\bm{h}^{l}_{e_j} - 2\bm{h}_{r_k}^T\bm{h}^{l}_{e_j}\bm{h}_{r_k})\right)
\end{equation}
here we employ $tanh$ as the activation function.
Instead of standard linear transformation matrix $\bm{W}$, we utilize \emph{Relational Projection} operation \cite{DBLP:conf/cikm/MaoWXWL20}.
Such operation generates relation-specific embedding for each entity without extra parameters.
As for the calculation of $\alpha_{ijk}$, we adopt the meta-path \cite{DBLP:conf/nips/YunJKKK19} mechanism to assign weights:
\begin{equation}
  \alpha^{l}_{ijk} = \frac{exp(\bm{v}^T\bm{h_{r_k}})}{\sum_{e_j'\in \mathcal{N}_{e_i}} \sum_{r_{k'} \in \mathcal{R}_{ij'}} exp(\bm{v}^T\bm{h}_{r_{k'}})}
\end{equation}
where $\bm{v}^T$ is an attention vector.
\emph{Softmax} operation selects the most critical path from all types of edges connected to the entities (i.e., meta-path), which embeds the relational anisotropy but simplifies the calculation to the greatest extent.

In previous studies \cite{DBLP:conf/aaai/SunW0CDZQ20,DBLP:conf/wsdm/MaoWXLW20}, GNN is able to expand to multi-hop neighboring level information by stacking more layers, thus to create a more global-aware representation of the graph.
Following this idea, we concatenate the embeddings from different layers together to obtain the \emph{Multi-Hop Embeddings} for entity $e_i$:
\begin{equation}
  \bm{h}^{multi}_{e_i} = [\bm{h}^0_{e_i}\|\bm{h}^1_{e_i}\|...\|\bm{h}^l_{e_i}]
\end{equation}
where $\|$ represents the concatenate operation.

\begin{figure}[t]
  \centering
  \includegraphics[width=0.9\linewidth]{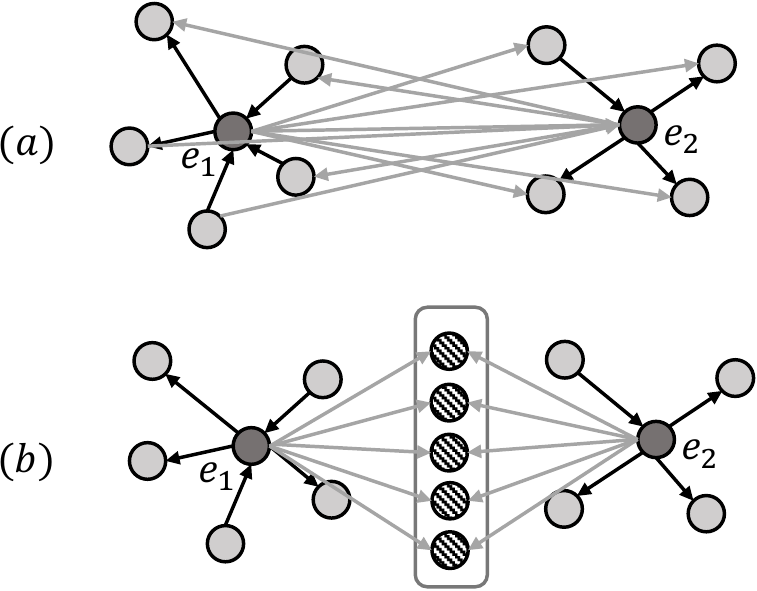}
  \caption{The illustration of two approaches. (a) represents \emph{Graph Matching Network} (GMN) and (b) represents \emph{Proxy Matching Attention Layer}.}\label{Figure:GMN}
\end{figure}

\subsection{Proxy Matching Attention Layer}
So far, the GNN encoders we have discussed only focus on modeling a single KG while leaving the cross-graph information to be learned by the alignment module alone.
Graph Matching Network \cite{DBLP:conf/icml/LiGDVK19} (GMN) builds a cross-graph attention mechanism to learn similarities, although they view the alignment purely as a node-to-node interaction (as illustrated in Figure \ref{Figure:GMN}(a)).
Formally, GMN measures the difference between $e_i \in E_1$ and its closest neighbor in the other graph as follows:

\begin{equation}
  \beta_{ij} = \frac{exp(sim(\bm{h}_{e_i},\bm{h}_{e_j}))} {\sum_{e_k \in E_2} exp(sim(\bm{h}_{e_i},\bm{h}_{e_k}))}
\end{equation}

\begin{equation}
  \bm{h}_{e_i}^{diff} = \sum_{e_j \in E_2} \beta_{ij}(\bm{h}_{e_i} - \bm{h}_{e_j})
\end{equation}
$sim(*)$ is a vector space similarity metric, $\bm{h}_{e_i}^{diff}$ represents the difference of $e_i$ against all entities from $G_2$.
Such node-to-node interaction enforces the embeddings to be learned jointly on a pair, at the cost of massive extra computation efficiency.
Since attention weights are required for every pair of nodes across two graphs, this operation has a computation cost of $O(|E_1||E_2|)$.
As mentioned in Section \ref{sec:intro}, GM-Align which incorporates GMN needs several days to obtain the results on the large-scale dataset (DWY$100$K).
Driven by similar motivation, but in our interpretation, alignment itself is nothing but a special relation type whose representation can be explicitly learned in early stage.

Inspired by the above, we propose the \emph{Proxy Matching Attention Layer}.
As shown in Figure \ref{Figure:GMN} (b), we employ a limited set of proxy vectors to represent the cross-graph alignment relation, similar to use anchor points to present a space.
If two entities are equivalent, their similarity distributions associated with these proxy vectors should also be consistent.
In this way, the proposed layer is able to capture the cross-graph alignment information without computing node-to-node interaction.
The interaction of the \emph{Proxy Matching Attention Layer} is to calculate the similarity between all entities and limited anchors, which is similar to clustering.
On large-scale KGs or dense graphs, this interaction approach can greatly reduce the computational complexity from O($|E_1||E_2|$) to O($|E_1|+|E_2|$).

The inputs of the \emph{Proxy Matching Attention Layer} are two matrices: $\bm{H}^{multi} \in \mathbb{R}^{|E| \times ld}$ represents the entities embeddings obtained by the \emph{Simplified Relational Attention Layer} and $\bm{Q} \in \mathbb{R}^{n \times ld}$ represents proxy vectors with random initialization, where $n$ represents the number of proxy vectors.
Just like GMN, the first step is to compute the similarity between each entity and all proxy vectors:
\begin{equation}
  \beta_{ij} = \frac{exp(cos(\bm{h}^{multi}_{e_i},\bm{q}_j))} {\sum_{k \in S_{p}} exp(cos(\bm{h}_{e_i},\bm{q}_{k}))}
\end{equation}
$S_{p}$ represents the set of proxy vectors.
Here we use the cosine metric to measure the similarity between embeddings.
Then, the cross-graph embedding for entity $e_i$ can be computed as:
\begin{equation}
  \bm{h}_{e_i}^{p} = \sum_{j \in S_{p}} \beta_{ij}(\bm{h}^{multi}_{e_i} - \bm{q}_{j})
\end{equation}
$\bm{h}_{e_i}^{p}$ intuitively describes the difference between $\bm{h}^{multi}_{e_i}$ and all proxy vectors.
Finally, we employ a gate mechanism \cite{DBLP:journals/corr/SrivastavaGS15} to combine $\bm{h}^{multi}_{e_i}$ and $\bm{h}^{p}_{e_i}$, controlling the information flow between single graph and multiple graphs:

\begin{equation}
  \bm{\eta}_{e_i} = sigmoid(\bm{M}\bm{h}^{p}_{e_i}+\bm{b})
\end{equation}

\begin{equation}
  \bm{h}_{e_i}^{final} = \bm{\eta}_{e_i}\cdot \bm{h}^{p}_{e_i} + (1-\bm{\eta}_{e_i})\cdot \bm{h}^{p}_{e_i}
\end{equation}
$\bm{M}$ and $\bm{b}$ are the gate weight matrix and gate bias vector.

\section{Normalized Hard Sample Mining}
Typically, in KGs, only a small portion of cross-graph entity pairs are aligned.
So negative sampling is crucial to EA methods.
However, the most common approach which selects the K-nearest neighbors, spends a lot of time on candidate ranking in each epoch.
In this section, we propose a \emph{Normalized Hard Sampling Mining} strategy, which is efficient and reduces loss shift.

\subsection{Smooth Hard Sample Mining}
Both TransE-based and GNN-based EA methods rely on the pair-wise loss functions to optimize the similarities between samples.
TransE-based methods use TransE loss to encode KGs:
\begin{equation}
  L = \sum_{(h,r,t)\in T} \left[\gamma + \|\bm{h}+\bm{r}-\bm{t}\|^2_2 - \|\bm{h}'+\bm{r}'-\bm{t}'\|^2_2\right]_{+}
\end{equation}
GNN-based methods use Triplet loss to map the embeddings from two KGs into a unified space:
\begin{equation}
  L = \sum_{(e_i,e_j)\in p} \left[\gamma + sim(e_i,e_j) - sim(e_i',e_j')\right]_{+}
\end{equation}
where $\gamma$ represents a fixed margin, $[x]_+$ represents the operation Max$(0,x)$, $x'$ represents the negative sample of $x$.

Initially, the negative samples in pair-wise loss are generated through uniform random sampling, but this kind of samples is highly redundant and comprises too many easy even uninformative samples.
Training with such low-quality negative samples may significantly degrade the model's learning capability and slow down the convergence.
A simple but effective strategy is to select the K-nearest neighbors around the positive sample to be negative samples.
This is also known as \textbf{Hard Sample Mining}.
BootEA proposes the \emph{Truncated Uniform Negative Sampling} (TUNS) based on this strategy and reports that it could significantly reduce the number of training epochs and improve performance.
Most of the subsequent works follow this approach, such as KECG \cite{DBLP:conf/emnlp/LiCHSLC19}, MuGNN \cite{DBLP:conf/acl/CaoLLLLC19}, TransEdge \cite{DBLP:conf/semweb/SunHL17}, and etc.
However, faster convergence cannot shorten the overall training time.
Because it has to spent massive time in candidate ranking for the next epoch, and this process is difficult to be fully parallelized on GPU.

In the field of deep metric learning, some studies \cite{DBLP:conf/cvpr/SunCZZZWW20,DBLP:conf/cvpr/SongXJS16} propose to use the \emph{LogSumExp} operation to smoothly generate hard negative samples:
\begin{equation}
  L = log\left[1 + \sum_{i\in P} \sum_{j\in N} exp(\lambda(\gamma + s_i-s_j))\right]
\end{equation}
where $P$ represents the positive sample set of the anchor and $N$ represents the negative sample set.
$\lambda$ is a scale factor.
If $\lambda\rightarrow\infty$:

\begin{equation}
\begin{aligned}
  L &= \underset{\lambda\rightarrow\infty}{lim} \frac{1}{\lambda}log\left[1 + \sum_{i\in P} \sum_{j\in N} exp(\lambda(\gamma + s_i-s_j))\right]\\
    &= Max[\gamma+s_i-s_j]_+
\end{aligned}
\end{equation}

\emph{LogSumExp} is approximate to TUNS with K = $1$.
When $\lambda$ is set to an appropriate value, \emph{LogSumExp} could replace the K-nearest sampling strategy to generate high-quality negative samples, but with better computational efficiency (because this process could be fully parallelized on GPU).
More interestingly, when $\lambda = 1$, the loss function is equivalent to \emph{Softmax} with \emph{Cross-Entropy} loss.
This also indicates that the classification losses and the pair-wise losses are essentially two sides of the same coin.

\subsection{Loss Normalization}
Both TUNS and \emph{LogSumExp} face the same dilemma of how to select the proper value for their hyper-parameters.
In TUNS, the hyper-parameter is the number of nearest neighbors K.
A small K will lead to slow convergence in the initial training process, while an overlarge K makes the negative samples too "easy."
In \emph{LogSumExp} operation, the hyper-parameter is the scalar factor $\lambda$.
As illustrated in Table \ref{dilemma}, if $\lambda$ is set too large, the weights of samples are greatly affected by the random disturbance at the beginning of training.
For example, when $\lambda=100$, these five pair losses are closer to each other while their corresponding weights vary a lot.
In such a case, the model would tend to only focus on a few samples, slowing down the convergence.
On the other hand, if $\lambda$ is too small, it would be difficult for the model to pick up hard samples in the later stage, which causes model degradation.
For example, when $\lambda=5$, though $l_1$ is seven times larger than $l_2$, the weights difference is small.

\begin{table}
\begin{center}
\renewcommand\arraystretch{1.1}
\begin{tabular}{cccccc}
\toprule
&$l_1$&$l_2$&$l_3$&$l_4$&$l_5$\\
\toprule
pair loss&1.01&0.99&0.98&1.02&1.00\\
$\frac{\partial L}{\partial l_i} (\lambda = 100)$&0.111&0.015&0.005&0.826&0.041\\
\toprule
pair loss&0.07&0.01&0.02&0.02&0.02\\
$\frac{\partial L}{\partial l_i} (\lambda = 5)$&0.211&0.156&0.164&0.164&0.156\\
\bottomrule
\end{tabular}
\caption{Examples of the dilemma of selecting hyper-parameter $\lambda$, where $l$ represents the loss of a pair $\gamma + s_i - s_j$.} \label{dilemma}
\end{center}
\end{table}

Inspired by batch normalization \cite{DBLP:conf/icml/IoffeS15} which reduces the internal covariate shift, we propose to use a normalization step that fixes the mean and variance of sample losses and reduces the dependence on the scale of the hyper-parameter.
Our overall loss function is defined as follow:

\begin{equation}
\begin{aligned}
  L &= \sum_{(e_i,e_j)\in P} log \left[1 + \sum_{e_j'\in E_2} exp(\lambda l_n(e_i,e_j,e_j') + \tau)\right]\\
    &+ \sum_{(e_i,e_j)\in P} log \left[1 + \sum_{e_i'\in E_1} exp(\lambda l_n(e_j,e_i,e_i') + \tau )\right]\\
\end{aligned}
\end{equation}

$l_n(e_i,e_j,e_j')$ represents the normalized loss of the triple $(e_i,e_j,e_j')$.
$\tau$ and $\lambda^2$ represent the new mean and the new variance of normalized loss respectively.
$l_n(e_i,e_j,e_j')$ is defined as follow:
\begin{equation}
  l_n(e_i,e_j,e_j') = \frac{l_o(e_i,e_j,e_j') - \mu(e_i,e_j)}{\sqrt{\sigma^2(e_i,e_j)-\epsilon}}
\end{equation}
\begin{equation}
  l_o(e_i,e_j,e_j') = \gamma + sim(e_i,e_j) - sim(e_i,e_j')
\end{equation}
where $l_o(e_i,e_j,e_j')$ represents the original loss of the triple $(e_i,e_j,e_j')$, $\mu$ and $\sigma^2$ represent the mean and the variance of original loss, which are computed by:
\begin{equation}
  \mu(e_i,e_j) = \frac{1}{|E_2|}\sum_{e_j'\in E_2} l_o(e_i,e_j,e_j')
\end{equation}
\begin{equation}
  \sigma^2(e_i,e_j) = \frac{1}{|E_2|}\sum_{e_j'\in E_2} \left[l_o(e_i,e_j,e_j') - \mu(e_i,e_j)\right]^2
\end{equation}
The calculation process of $l_n(e_j,e_i,e_i')$ is similar to $l_n(e_i,e_j,e_j')$.

During training, we choose L$2$ distance as the metric to measure the similarity between entities:
\begin{equation}
  sim(e_i,e_j) = \|h^{final}_{e_i} - h^{final}_{e_j}\|^2_2
\end{equation}
During testing, in order to address the hubness problem in high-dimensional space, CSLS \cite{DBLP:conf/iclr/LampleCRDJ18} is set to be the distance metric.
Note that in training, $\sigma$ and $\mu$ won't participate in gradient calculation and backpropagation.
This is because our loss normalization is designed to change the weights of the samples, not the gradient direction.
If $\sigma$ and $\mu$ are updated in the backpropagation step, our loss will fail to converge.

\section{Experiments}
We use the Keras framework for developing our approach.
Our experiments are conducted on a workstation with a GeForce GTX TITAN X GPU and $128$GB memory, which is consistent with the summary \cite{9174835}.
The code is now available on GitHub \footnote{https://github.com/MaoXinn/Dual-AMN}.

\begin{table}[t]
\begin{center}
\resizebox{0.8\linewidth}{!}{
\renewcommand\arraystretch{1.1}
\begin{tabular}{p{1.55cm}c|cccccc}
\toprule
\multicolumn{2}{c|}{Datasets} & $|E|$ & $|R|$  & $|T|$\\
\toprule
\multirow{2}{1.3cm}{$\rm{DWY_{DBP-YG}}$} & DBpedia & 100,000 &  302  & 428,952\\
& YAGO3 & 100,000 & 31  &  502,563 \\
\multirow{2}{1.3cm}{$\rm{DWY_{DBP-WD}}$} & DBpedia & 100,000 &  330 & 463,294\\
& Wikipedia & 100,000 &  220  &  448,774  \\
\hline
\multirow{2}{1.3cm}{$\rm{DBP_{ZH-EN}}$} & Chinese & 19,388 & 1,701& 70,414\\
& English & 19,572 & 1,323 & 95,142 \\
\multirow{2}{1.3cm}{$\rm{DBP_{JA-EN}}$} & Japanese & 19,814 & 1,299 & 77,214\\
& English & 19,780 & 1,153  & 93,484 \\
\multirow{2}{1.3cm}{$\rm{DBP_{FR-EN}}$} & French & 19,661 & 903 & 105,998\\
& English & 19,993 & 1,208 & 115,722  \\
\hline
\multirow{2}{1.3cm}{$\rm{SRPRS_{FR-EN}}$} & French & 15,000 & 177& 33,532\\
& English & 15,000 & 221& 36,508 \\
\multirow{2}{1.3cm}{$\rm{SRPRS_{DE-EN}}$} & German & 15,000 & 120 & 37,377\\
& English & 15,000 & 222 & 38,363  \\
\multirow{2}{1.3cm}{$\rm{SRPRS_{DBP-YG}}$} & DBpedia & 15,000 &  223  & 33,748\\
& YAGO3 & 15,000 & 30  &  36,569 \\
\multirow{2}{1.3cm}{$\rm{SRPRS_{DBP-WD}}$} & DBpedia & 15,000 &  253 & 38,421\\
& Wikipedia & 15,000 & 144 &  40,159  \\
\bottomrule
\end{tabular}
}
\end{center}
\caption{Statistical data of DBP15K, DWY100K and SRPRS.}\label{table:data}
\end{table}

\subsection{Datasets}
To fairly and comprehensively verify the effectiveness, robustness and scalability of our model, we construct experiments on three widely used public datasets:

(1) \textbf{DBP15K} \cite{DBLP:conf/semweb/SunHL17}:
This dataset consists of three cross-lingual subsets constructed from DBpedia: English-French ($\rm DBP_{EN-FR}$), English-Chinese ($\rm DBP_{EN-ZH}$), English-Japanese ($\rm DBP_{EN-JA}$).
Each subset contains $15,000$ pre-aligned entity pairs for training and testing.
As an early dataset, DBP$15$K is widely used but has some defects: small scale and dense links.
These defects prompt more datasets to be proposed.

(2) \textbf{DWY100K} \cite{DBLP:conf/ijcai/SunHZQ18}:  This dataset comprises two mono-lingual subsets, each containing $100,000$ pre-aligned entities pairs and nearly one million triples.
$\rm DWY_{DBP-WD}$ represents the subset extracted from DBpedia and Wikidata, and $\rm DWY_{DBP-YG}$ represents DBpedia and YAGO.
$\rm DWY100K$, as the largest dataset of the three, brings challenges to space and time complexity.

(3) \textbf{SRPRS} \cite{DBLP:conf/icml/GuoSH19}:
Compared with the real-world KGs, the above two datasets are too dense, and the degree distribution is quite different from the real.
Thus, \citet{DBLP:conf/icml/GuoSH19} propose a sparse dataset, including two cross-lingual subsets ($\rm SRPRS_{FR-EN}$ and $\rm SRPRS_{DE-EN} $) and two mono-lingual subsets ($\rm SRPRS_{DBP-WD}$ and $\rm SRPRS_{DBP-WD}$).
Same with DBP$15$K, each subset of SRPRS contains $15,000$ pre-aligned entity pairs for training and testing.
This dataset challenges the modeling ability of EA approaches when facing limited information.

The statistics of these datasets are listed in Table \ref{table:data}.
Consistent with previous studies, we randomly split $30\%$ of the pre-aligned entity pairs for training and developing, while the remaining $70\%$ for testing.

\begin{table*}[!t]
\begin{center}
\resizebox{\textwidth}{!}{
\renewcommand\arraystretch{1.1}
\begin{tabular}{c|cccccccccccccccc}
  \toprule
  \multicolumn{2}{c}{\multirow{2}{*}{Method}} & \multicolumn{3}{c}{$\rm{DBP_{ZH-EN}}$} & \multicolumn{3}{c}{$\rm{DBP_{JA-EN}}$} & \multicolumn{3}{c}{$\rm{DBP_{FR-EN}}$}& \multicolumn{3}{c}{$\rm{SRPRS_{FR-EN}}$}& \multicolumn{3}{c}{$\rm{SRPRS_{DE-EN}}$}  \\
  \multicolumn{2}{c}{} & H@1 & H@10 & MRR & H@1 & H@10 & MRR & H@1 & H@10 & MRR & H@1 & H@10 & MRR & H@1 & H@10 & MRR\\
  \hline
  \multirow{6}*{Basic}
  & MTransE & 0.209 & 0.512 & 0.310 & 0.250 & 0.572 & 0.360 & 0.247 & 0.577 & 0.360 &0.213& 0.447&0.290&0.107&0.248&0.160\\
  & GCN-Align & 0.434 & 0.762 & 0.550 & 0.427 & 0.762 & 0.540 & 0.411 & 0.772 & 0.530 & 0.243& 0.522&0.340&0.385&0.600& 0.460\\
  & MuGNN & 0.494 & 0.844 & 0.611 & 0.501 & 0.857 & 0.621 & 0.495 & 0.870  &0.621 &0.131&0.342& 0.208& 0.245&0.431&0.310\\
  & KECG&0.477&0.835&0.598& 0.489&0.844& 0.610&0.486&0.851& 0.610& 0.298& 0.616& 0.403&0.444 &0.707& 0.540\\
  & RSNs&0.508&0.745&0.591&0.507&0.737&0.590&0.516&0.768&0.605&0.350&0.636& 0.440& 0.484& 0.729&0.570\\
  & \textbf{Dual-AMN} &\textbf{0.731}&\textbf{0.923}&\textbf{0.799}&\textbf{0.726}&\textbf{0.927}&\textbf{0.799}&\textbf{0.756}&\textbf{0.948}&\textbf{0.827}&\textbf{0.452}&\textbf{0.748}&\textbf{0.552}&\textbf{0.591}&\textbf{0.820}&\textbf{0.670}\\
  \hline
  \multirow{5}*{Semi}
  & BootEA & 0.629 & 0.847 & 0.703 & 0.622 & 0.853 & 0.701 & 0.653 & 0.874 & 0.731 & 0.365&0.649&0.460&0.503&0.732&0.580\\
  & NAEA & 0.650 & 0.867 & 0.720 & 0.641 & 0.872 & 0.718 & 0.673 & 0.894 & 0.752  &0.177&0.416& 0.260&0.307&0.535&0.390\\
  & TransEdge&0.735&0.919&0.801&0.719&0.932&0.795&0.710&0.941&0.796&0.400&0.675&0.490&0.556&0.753&0.630\\
  & MRAEA &0.757&0.930&0.827&0.758&0.934&0.826&0.781&0.948&0.849&0.460&0.768&0.559&0.594&0.818&0.666\\
  & \textbf{Dual-AMN} &\textbf{0.808}&\textbf{0.940}&\textbf{0.857}&\textbf{0.801}&\textbf{0.949}&\textbf{0.855}&\textbf{0.840}&\textbf{0.965}&\textbf{0.888}&\textbf{0.481}&\textbf{0.778}&\textbf{0.568}&\textbf{0.614}&\textbf{0.823}&\textbf{0.687}\\
  \hline
  \multirow{5}*{Literal}
  & GM-Align & 0.679 & 0.785 & - & 0.739 & 0.872 & - & 0.894 & 0.952 & - & 0.574&0.646&0.602&0.681&0.748&0.710\\
  & RDGCN & 0.697 & 0.842 & 0.750 & 0.763 & 0.897 & 0.810 & 0.873 & 0.950 & 0.901  &0.672&0.767& 0.710&0.779&0.886&0.820 \\
  & HMAN&0.561&0.859&0.670&0.557&0.860&0.670&0.550&0.876&0.660&0.401&0.705&0.500&0.528&0.778&0.620\\
  & HGCN &0.720&0.857&0.760&0.766&0.897&0.810&0.892&0.961&0.910&0.670&0.770&0.710&0.763&0.863&0.801\\
  & \textbf{Dual-AMN} &\textbf{0.861}&\textbf{0.964}&\textbf{0.901}&\textbf{0.892}&\textbf{0.978}&\textbf{0.925}&\textbf{0.954}&\textbf{0.994}&\textbf{0.970}&\textbf{0.802}&\textbf{0.932}&\textbf{0.851}&\textbf{0.891}&\textbf{0.972}&\textbf{0.923}\\
    \bottomrule
\end{tabular}
}
\caption{Experimental results on cross-lingual datasets.}
\label{table:res1}
\resizebox{\textwidth}{!}{
\renewcommand\arraystretch{1}
\begin{tabular}{c|ccccccccccccc}
  \toprule
  \multicolumn{2}{c}{\multirow{2}{*}{Method}} & \multicolumn{3}{c}{$\rm{DWY_{DBP-WD}}$}& \multicolumn{3}{c}{$\rm{DWY_{DBP-YG}}$}& \multicolumn{3}{c}{$\rm{SRPRS_{DBP-WD}}$}& \multicolumn{3}{c}{$\rm{SRPRS_{DBP-YG}}$}\\
  \multicolumn{2}{c}{} & H@1 & H@10 & MRR & H@1 & H@10 & MRR & H@1 & H@10 & MRR & H@1 & H@10 & MRR \\
  \hline
  \multirow{6}*{Basic}
  & MTransE & 0.238 & 0.507 & 0.330 & 0.227 & 0.414 & 0.290 & 0.188 & 0.382 & 0.260 &0.196& 0.401&0.270\\
  & GCN-Align & 0.494 & 0.756 & 0.590 & 0.598 & 0.829 & 0.680 & 0.291 & 0.556 & 0.380 & 0.319& 0.586&0.410\\
  & MuGNN & 0.604 & 0.894 & 0.701 & 0.739 & 0.937 & 0.810 & 0.151 & 0.366  &0.220 &0.175&0.381& 0.240\\
  & KECG&0.631&0.888&0.720& 0.719&0.904& 0.790&0.323&0.646& 0.430& 0.350& 0.651& 0.450\\
  & RSNs&0.607&0.793&0.673&0.689&0.878&0.756&0.391&0.663&0.480&0.393&0.665& 0.490\\
  & \textbf{Dual-AMN} &\textbf{0.786}&\textbf{0.952}&\textbf{0.848}&\textbf{0.866}&\textbf{0.977}&\textbf{0.907}&\textbf{0.513}&\textbf{0.801}&\textbf{0.609}&\textbf{0.495}&\textbf{0.790}&\textbf{0.596}\\
  \hline
  \multirow{5}*{Semi}
  & BootEA & 0.748 & 0.898 & 0.801 & 0.761 & 0.894 & 0.808 & 0.384 & 0.667 & 0.480 & 0.381&0.651&0.470\\
  & NAEA & - & - & - & - & - & - & 0.182 & 0.429 & 0.260  &0.195&0.451& 0.280\\
  & TransEdge&0.788&0.938&0.824&0.792&0.936&0.832&0.461&0.738&0.560&0.443&0.699&0.530\\
  & MRAEA &0.794&0.930&0.856&0.819&0.951&0.875&0.509&0.795&0.597&0.485&0.768&0.574\\
  & \textbf{Dual-AMN} &\textbf{0.869}&\textbf{0.969}&\textbf{0.908}&\textbf{0.907}&\textbf{0.981}&\textbf{0.935}&\textbf{0.546}&\textbf{0.813}&\textbf{0.635}&\textbf{0.518}&\textbf{0.795}&\textbf{0.613}\\
    \bottomrule
\end{tabular}
}
\caption{Experimental results on mono-lingual datasets.
Because of the memory limitation, NAEA cannot work on DWY$100$K.
}
\label{table:res2}
\end{center}
\end{table*}

\subsection{Baselines}
As mentioned in Section \ref{sec:rw}, many studies adopt enhancement modules.
For instance, GM-Align and RDGCN propose to introduce literal information to provide a multi-aspect view.
The introduction of additional information leads to unfair comparisons between methods.
Thus, existing EA methods will be compared separately according to the enhancement category:

(1) \textbf{Basic}:
This kind of method only uses the original structure information (i.e., triples) in the dataset and does not introduce any extra enhancement module:
MTransE \cite{DBLP:conf/ijcai/ChenTYZ17}, GCN-Align \cite{DBLP:conf/emnlp/WangLLZ18}, RSNs \cite{DBLP:conf/icml/GuoSH19}, MuGNN \cite{DBLP:conf/acl/CaoLLLLC19}, KECG \cite{DBLP:conf/emnlp/LiCHSLC19}.

(2) \textbf{Semi-supervised}:
These methods adopt bootstrapping to generate semi-supervised structure data:
BootEA \cite{DBLP:conf/ijcai/SunHZQ18}, NAEA \cite{DBLP:conf/ijcai/ZhuZ0TG19}, TransEdge \cite{DBLP:journals/corr/abs-2004-13579}, and MRAEA \cite{DBLP:conf/wsdm/MaoWXLW20}.

(3) \textbf{Literal}:
To obtain a multi-aspect view, literal methods use literal information (e.g., entity name) of entities as input features:
GM-Align \cite{DBLP:conf/acl/XuWYFSWY19}, RDGCN \cite{DBLP:conf/ijcai/WuLF0Y019}, HMAN \cite{DBLP:conf/emnlp/YangZSLLS19}, HGCN \cite{DBLP:conf/emnlp/WuLFWZ19}.

To make a fair comparison against above three types of methods, our model also has three corresponding versions:
(1) Dual-AMN is the basic version without any enhancement module, as described in Section \ref{sec:model}.
(2) Dual-AMN (Semi) introduces the bi-directional iterative strategy proposed by MRAEA to generate semi-supervised data.
(3) Dual-AMN (Lit) adopts a simple strategy to utilize literal information.
For $e_i \in KG_1$ and $e_j \in KG_2$, we use Dual-AMN (Semi) to obtain the structural similarity $s_{ij}$.
Then, using the cross-lingual word embedding \footnote{Same with GM-Align \cite{DBLP:conf/iclr/LampleCRDJ18}.} to calculate the literal similarity $l_{ij}$.
Finally, the entities are ranked according to $l_{ij} + s_{ij}$.

\subsection{Experimental Settings}
\textbf{Metrics}.
Following convention, we use $Hits@k$ and \emph{Mean Reciprocal Rank} ($MRR$) as our evaluation metrics.
The $Hits@k$ score is calculated by measuring the proportion of correctly aligned pairs in the top-k.
In particular, $Hits@1$ represents accuracy.
In order to be convincing, the reported performance is the average of five independent training runs.

\noindent
\textbf{Hyper-parameters}.
For all dataset, we use a same config:
The dimensionality for embeddings $d = 100$;
depth of GNN $l = 2$;
number of proxy vectors $n = 64$;
margin $\gamma = 1$;
new mean and new variance of normalized loss are $\tau = 10$ and $\lambda = 30$;
batch size is $1024$; dropout rate is set to $30\%$.
RMSprop is adopted to optimize the model with learning rate set to $0.005$.

\subsection{Main Experiments}
In Table \ref{table:res1} and Table \ref{table:res2}, we report the performances of all methods on cross-lingual datasets and mono-lingual datasets, respectively.
We compare the performances within each category.

\noindent
\textbf{Dual-AMN vs. Basic Methods}.
Our method consistently achieves the best performance across all datasets.
On the small-scale dense dataset (DBP$15$K), Dual-AMN outperforms other methods by at least $20\%$ in terms of both $Hits@1$ and $MRR$.
On the large-scale dense dataset (DWY$100$K), the performances are increased by more than $15\%$ compared to previous SOTA.
Experimental results show that the designs of Dual-AMN effectively captures the rich structural information of these two datasets.
By cutting down the number of triples, SRPRS challenges EA methods' ability to model sparse KGs.
It is not surprising to see that the performances of all methods drop significantly compared to the results on dense datasets.
RSNs outperforms the previous SOTA on this dataset, which could be credited to the long-term relational paths it captures.
But our Dual-AMN still achieves the best performance, exceeding RSNs by at least $10\%$ on $Hits@1$ and $MRR$.
All these experimental results demonstrate the effectiveness of Dual-AMN in capturing the structural information.

\noindent
\textbf{Dual-AMN vs. Semi-supervised Methods}.
Benefiting from the semi-supervised strategy to generate more labeled data for the next training round, the overall performances of the semi-supervised methods surpass the basic methods.
Compared with previous SOTA methods, our method outperforms at least $5\%$ on $Hits@1$.
Compared to its own basic version, the semi-supervised strategy greatly improves the performances on DBP$15$K and DWY$100$K.
On SRPRS, although the semi-supervised strategy still has some benefit, the improvement is reduced to $2\%\sim3\%$.
We believe the reason for the smaller improvement is because the sparse nature of SRPRS makes its structure information insufficient to generate high-quality semi-supervised data.
Overall, the semi-supervised strategy performs well on dense datasets, while its improvement is marginal in sparse datasets.

\noindent
\textbf{Dual-AMN vs. Literal Methods}.
According to \citet{9174835}, because the entity names between mono-lingual KGs are almost identical, the edit distance algorithm could achieve the ground-truth performance.
Therefore, the literal methods only experiment on cross-lingual datasets.

By combining with cross-lingual embeddings, the performances of Dual-AMN are further improved and surpass the previous SOTA methods across all datasets.
From observing Table \ref{table:res1}, we found that the performances of the literal methods vary significantly according to language pairs, which is completely different from the structure-only methods.
On DBP$15$K, the introduction of literal information increases $Hits@1$ by $6\%$, $9\%$, and $11\%$, which indicates that French is the most similar language to English, while Chinese is the most different.
Besides, due to the lack of structural information, the literal information is more critical on SRPRS.
Literal information improves the performances by $30\%$ on $Hits@1$.

It must be admitted that our way of utilizing literal information is too simple and crude.
Compared with other methods, performance improvement mainly comes from better structural embeddings.
How to better integrate literal information is our future work.

\begin{table}
\begin{center}
\resizebox{0.9\linewidth}{!}{
\renewcommand\arraystretch{0.9}
\begin{tabular}{cccc}
  \toprule
  \textbf{Method}&\textbf{DBP15K}&\textbf{SRPRS}&\textbf{DWY100K}\\
  \toprule
  MTransE \cite{DBLP:conf/ijcai/ChenTYZ17} &6,467&3,355&70,085\\
  GCN-Align \cite{DBLP:conf/emnlp/WangLLZ18}&103&87&3,212\\
  RSNs \cite{DBLP:conf/icml/GuoSH19}&7,539&2,602&28,516\\
  MuGNN \cite{DBLP:conf/acl/CaoLLLLC19}&3,156&2,215&47,735\\
  KECG \cite{DBLP:conf/emnlp/LiCHSLC19}&3,724&1,800&125,386\\
  \textbf{Dual-AMN} \cite{DBLP:conf/aaai/SunW0CDZQ20}&\textbf{35}&\textbf{27}&\textbf{1,094}\\
  \midrule
  BootEA \cite{DBLP:conf/ijcai/SunHZQ18}&4,661&2,659&64,471\\
  NAEA \cite{DBLP:conf/ijcai/ZhuZ0TG19}&19,115&11,746&-\\
  TransEdge\cite{DBLP:journals/corr/abs-2004-13579}&3,629&1,210&20,839\\
  MRAEA \cite{DBLP:conf/wsdm/MaoWXLW20}&3,894&1,248&23,275\\
  \textbf{Dual-AMN(Semi)} \cite{DBLP:conf/aaai/SunW0CDZQ20}&\textbf{85}&\textbf{79}&\textbf{3,169}\\
  \midrule
  GM-Align \cite{DBLP:conf/acl/XuWYFSWY19}&26,328&13,032&459,715\\
  RDGCN \cite{DBLP:conf/ijcai/WuLF0Y019}&6,711&886&-\\
  HMAN \cite{DBLP:conf/emnlp/YangZSLLS19}&5,455&4,424&31,895\\
  HGCN \cite{DBLP:conf/emnlp/WuLFWZ19}&11,275&2,504&60,005\\
  \textbf{Dual-AMN(Lit)} \cite{DBLP:conf/aaai/SunW0CDZQ20}&\textbf{101}&\textbf{96}&\textbf{3,257}\\
  \bottomrule
\end{tabular}
}
\end{center}
\caption{Time costs of EA methods (seconds).}\label{tabel:time}
\end{table}

\noindent
\textbf{Efficiency Analysis}.
Better performance is just the cherry on the cake.
Dual-AMN's trump card is superior efficiency.
Table \ref{tabel:time} reports the overall time costs of existing EA methods on each dataset, including data loading, pre-processing, training, and evaluating.
All results are obtained by directly running the source code provided by the authors.
And hyper-parameters are set to be the same as reported in their original papers.
Certainly, implement details such as learning rate, batch size, and pre-processing might influence the time costs.
However, we believe that these experimental results still reflect the overall efficiency of EA methods.

Obviously, the efficiency of Dual-AMN far exceeds competitors.
The time costs of complex EA methods are tens or even hundreds of times more than that of Dual-AMN.
Even compared with the fastest baseline (i.e., GCN-Align), the speed of Dual-AMN is $3\times$ faster, while the $Hits@1$ outperforms more than $20\%$.
Comparing Dual-AMN and Dual-AMN (Semi), semi-supervised strategy increases the time consumption about three times.
Due to the simple combining strategy, Dual-AMN (Lit) hardly increases the time consumption.

In particular, the large-scale dense dataset (DWY$100$K) poses a severe challenge to the space and time complexity of all EA methods.
Due to the limitation of GPU memory, MuGNN, KECG, and HMAN have to be run on CPU, resulting in massive time costs.
GM-Align is the least efficient method, because it uses GMN and requires a complicated pre-processing.
We fail to obtain results for NAEA and RDGCN in our experiment environment because they require extremely high memory space.
Benefit from the simplification of the encoder architecture and the \emph{Normalized Hard Sample Mining Loss}, our model could fully utilize the GPU to obtain high-accuracy results efficiently.
Even using the semi-supervised strategy for data augmentation, the proposed method still could obtain results within an hour.

In summary, the high efficiency of Dual-AMN makes the entity alignment application on large-scale KGs possible.

\begin{table}[t]
\renewcommand\arraystretch{1.5}
\centering
\resizebox{1.0\linewidth}{!}{
\begin{tabular}{l|cccccc}
\toprule
\multirow{2}{* }{Method} & \multicolumn{2}{c}{$\rm{DBP_{ZH-EN}}$} & \multicolumn{2}{c}{$\rm{DBP_{JA-EN}}$} & \multicolumn{2}{c}{$\rm{DBP_{FR-EN}}$}\\
& Hits@1 & MRR & Hits@1 & MRR & Hits@1 & MRR\\
\toprule
 Dual-AMN &$.731_{\pm .002}$&$.799_{\pm .002}$&$.726_{\pm .003}$&$.799_{\pm .002}$&$.756_{\pm .004}$&$ .827_{\pm .005}$\\
 \;\;-RA.&$.705_{\pm .001}$&$.781_{\pm .003}$&$.706_{\pm .002}$&$.785_{\pm .004}$&$.741_{\pm .004}$&$ .817_{\pm .002}$\\
 \;\;-RP.&$.702_{\pm .002}$&$.779_{\pm .002}$&$.704_{\pm .005}$&$.783_{\pm .003}$&$.745_{\pm .004}$&$ .821_{\pm .003}$\\
 \;\;-MHE. &$.656_{\pm .003}$&$.743_{\pm .002}$&$.658_{\pm .002}$&$.748_{\pm .001}$&$.698_{\pm .004}$&$ .783_{\pm .003}$\\
 \;\;-PAM. &$.711_{\pm .002}$&$.785_{\pm .001}$&$.710_{\pm .001}$&$.783_{\pm .002}$&$.738_{\pm .002}$&$ .812_{\pm .001}$\\
\bottomrule
\end{tabular}
}
\caption{Ablation experiment of architecture on DBP15K.}
\label{table:model_ablation}
\end{table}

\subsection{Ablation Experiment}
To demonstrate the effectiveness of each design in architecture and loss function, we construct two ablation experiments on DBP$15$K.

\noindent
\textbf{Ablation Experiment of architecture}.
Dual-AMN adopts the following four components to capture multi-aspect information existing in KGs:
(1) \emph{Relational Attention} mechanism (RA) finds the critical path around entities.
(2) \emph{Relational Projection} operation (RP) generates the relation-specific embedding for entities.
(3) \emph{Multi-Hop Embeddings} (MHE) creates a more global-aware representation of the KGs.
(4) \emph{Proxy Attention Matching Layer} (PAM) captures the cross-graph information.
Table \ref{table:model_ablation} reports the performances with $\rm Means_{\pm stds}$ after removing these components from Dual-AMN.
Among all these components, MHE has the greatest impact on performance.
Without MHE, the performance is degraded by at least $6\%$ on $Hits@1$.
Only stacking GNN layers cannot fully capture the global information, it is necessary to concatenate the output embeddings of each layer explicitly.
Besides, the remaining three components also show the necessity as our expectation.
On average, adopting these technologies improves performance by $2\%$ to $~3\%$.
By adopting these new designs, Dual-AMN further breaks the ceiling of EA accuracy.

\begin{figure}
  \centering
  \includegraphics[width=1\linewidth]{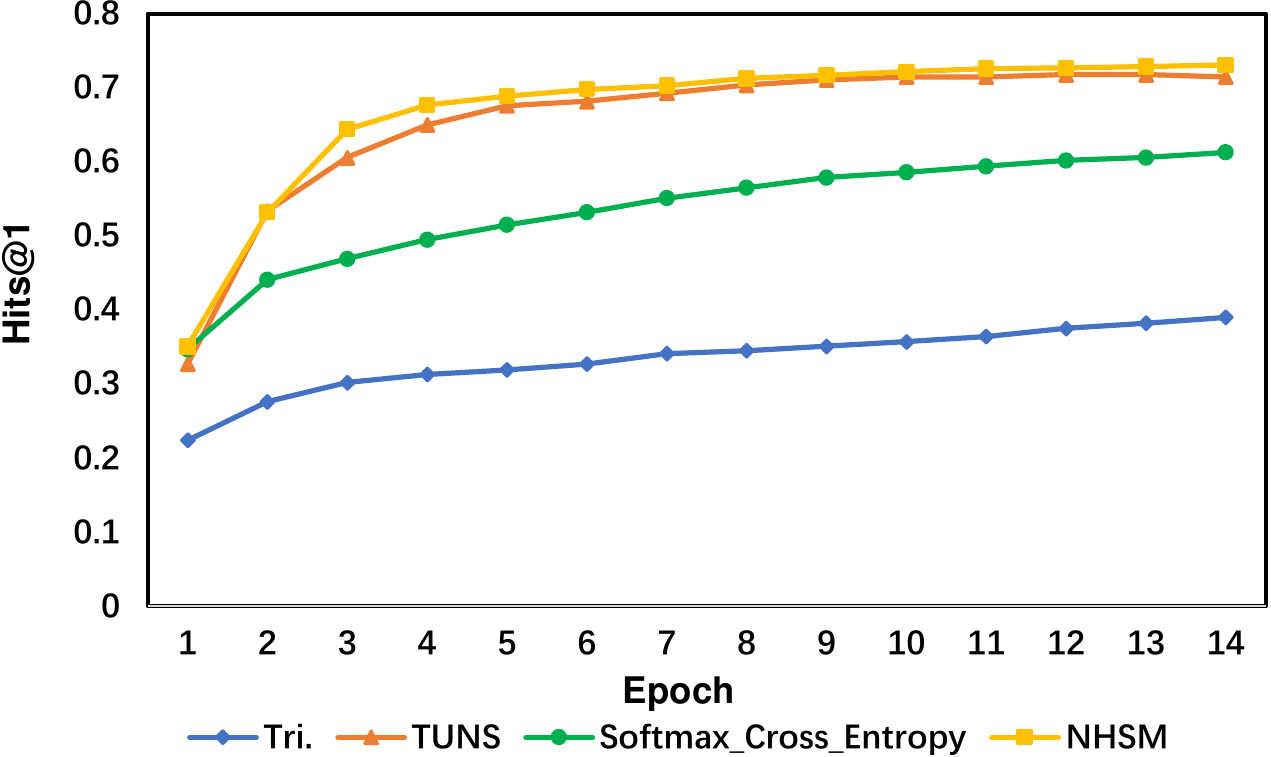}
  \caption{Ablation experiment of loss on $\rm DBP_{ZH-EN}$.
  Tri. represents Triplet Loss;
  TUNS represents Triplet loss with \emph{Truncated Uniform Negative Sampling Strategy.}
  For TUNS, top-$10$ nearest neighbors are selected as negative samples, following the same setting in most EA methods'.
  NHSM represents our \emph{Normalized Hard Sample Mining Loss.}}\label{Figure:loss}
\end{figure}
\noindent
\textbf{Ablation Experiment of Loss}.
Besides architecture, the \emph{Normalized Hard Sample Mining Loss} is also one of our main contributions.
To verify its effectiveness, we compare it with several common loss functions.
The results are visualized in Fig \ref{Figure:loss}.
Compared with the other three, the proposed loss could make the model converge faster and achieve the best performance.
\emph{Truncated Uniform Negative Sampling Strategy} also has a similar decent performance.
However, as we have mentioned, this sampling strategy requires massive time consumption.
Since most of the negative samples are redundant, the Triplet loss has the worst efficiency of all loss functions.
In our experiments, the Triplet loss function usually needs thousands of epochs to converge, and the performance is lower than the proposed loss about $4\%$.
The performance of \emph{Softmax} with \emph{Cross-Entropy} is stronger than Triplet loss, but there is obviously a performance gap with \emph{Normalized Hard Sample Mining Loss}.
These experimental results show that the proposed loss function significantly increases the convergence speed without losing any accuracy.
\subsection{Relation Interpretability}
\begin{table}[t]
\center
\resizebox{0.9\linewidth}{!}{
\renewcommand\arraystretch{1}
\begin{tabular}{c|c|c}
  \toprule
  Importance & $I_{r_k}$ & Examples \\
  \toprule
  \multirow{2}{*}{High} &\multirow{2}{*}{$[5,\infty)$}& $Parent$, $Starring$, $Capital$, $Father$, \\
        &&$County$, $Doctoral\_Advisor$ \\
  \hline
  \multirow{2}{*}{Medium} &\multirow{2}{*}{$[-5,5)$}& $Part$, $Inflow$, $Outflow$, $Tenant$, \\
        &&$Academic\_Advisor$ \\
  \hline
  \multirow{2}{*}{Low} &\multirow{2}{*}{$[-\infty,-5)$}& $President$, $Monarch$, $Bishop$, $Speaker$\\
        &&$Leader\_Party$, $Prime\_Minister$ \\
  \bottomrule
\end{tabular}
}
\caption{Relation examples of different importance.}
\label{sample}
\end{table}

In addition to the performance and speed advantages, our model also has a certain degree of interpretability.
Because the weights of adjacent entities are determined by the relations between them, thus these weights can reflect the importance of different relations to some extent.
The importance of each relation is obtained by the following equation:
\begin{equation}
  I_{r_k} = v^T h_{r_k}
\end{equation}
We train the model on the $\rm{DWY_{YG}}$ and output the importance $I_{r_k}$ of relations.
After clustering the relations according to $I_{r_k}$, we obtain the Table \ref{sample}.
From the observation, we summarize an interesting phenomenon.
The relations with high importance (i.e., meta-path) are usually able to identify the entity from another.
For example, if holding an $Person$ entity and $Parent$ relation, we can reduce the potential options down to a small space.
However, $President$ relation does not have this ability. A country can have many presidents, so its importance becomes extremely low.
Of course, this is inseparable from the characteristics of the $\rm{DBP_{YG}}$ dataset, which contains a large number of celebrities, especially the president, prime minister, and so on.
Therefore, such kinds of relations become unimportant in this dataset.

\begin{figure}
  \centering
  \includegraphics[width=1\linewidth]{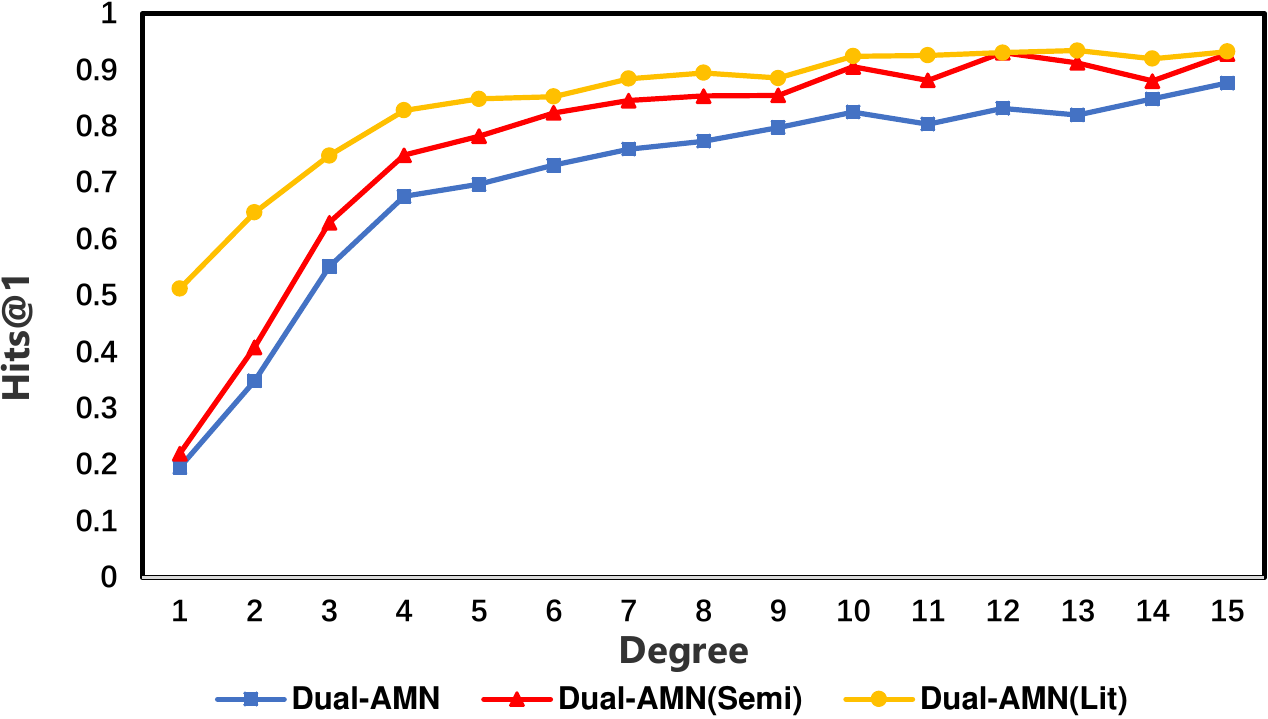}
  \caption{$Hits@1$ of the entities with different degrees.}\label{Figure:error}
\end{figure}

\subsection{Degree Analysis}
The main experiments show that the performances of all EA methods on sparse datasets are much lower than that of standard datasets.
In order to further explore the correlation between model performance and dataset density, we design an experiment on $\rm DBP_{ZH-EN}$.
Figure \ref{Figure:error} shows the $Hits@1$ of the three variants on different levels of entity degrees.
We observe a strong correlation between performance and degree.
As the degree increasing, the model performance improves significantly.
For Dual-AMN, the $Hits@1$ of the entities with one neighbor is only $20\%$.
The introduction of semi-supervised strategy improves the overall performance of the model, but it has a limited effect on those entities with extremely sparse local structures.
In sparse graphs, it is difficult to make correct inferences only based on limited structural information.
On the other hand, Dual-AMN (Lit) has much higher performance when the degree value is small, which proves that the incorporation of literal information effectively improves the accuracy of these sparse entities.
However, this strategy cannot work on the datasets without literal information.
Therefore, how to better represent these sparse entities without extra information is a key point of future work.

\section{Conclusion}
Over complex graph encoders and inefficient negative sampling strategies lead to the general inefficiency of existing EA methods, resulting in difficulty for applying on large-scale KGs.
In this paper, we propose a novel KG encoder \emph{Dual Attention Matching Network} (Dual-AMN), which not only models both intra-graph and cross-graph relations smartly but also greatly reduces computational complexity.
To replace the inefficient sampling strategy, we propose \emph{Normalized Hard Sample Mining Loss} to cut down the sampling consumption and accelerate the convergence speed.
These two modifications enable the proposed model to achieve the SOTA performance while the speed is several times than other EA methods.
The main experiments indicate that our method outperforms competitors across all datasets and metrics.
Furthermore, we design auxiliary experiments to demonstrate the effectiveness of each component and the interpretability of the model.

\bibliographystyle{ACM-Reference-Format}
\bibliography{sample-base}


\begin{thebibliography}{35}


\ifx \showCODEN    \undefined \def \showCODEN     #1{\unskip}     \fi
\ifx \showDOI      \undefined \def \showDOI       #1{#1}\fi
\ifx \showISBNx    \undefined \def \showISBNx     #1{\unskip}     \fi
\ifx \showISBNxiii \undefined \def \showISBNxiii  #1{\unskip}     \fi
\ifx \showISSN     \undefined \def \showISSN      #1{\unskip}     \fi
\ifx \showLCCN     \undefined \def \showLCCN      #1{\unskip}     \fi
\ifx \shownote     \undefined \def \shownote      #1{#1}          \fi
\ifx \showarticletitle \undefined \def \showarticletitle #1{#1}   \fi
\ifx \showURL      \undefined \def \showURL       {\relax}        \fi
\providecommand\bibfield[2]{#2}
\providecommand\bibinfo[2]{#2}
\providecommand\natexlab[1]{#1}
\providecommand\showeprint[2][]{arXiv:#2}

\bibitem[\protect\citeauthoryear{Auer, Bizer, Kobilarov, Lehmann, Cyganiak, and
  Ives}{Auer et~al\mbox{.}}{2007}]%
        {DBLP:conf/semweb/AuerBKLCI07}
\bibfield{author}{\bibinfo{person}{S{\"{o}}ren Auer},
  \bibinfo{person}{Christian Bizer}, \bibinfo{person}{Georgi Kobilarov},
  \bibinfo{person}{Jens Lehmann}, \bibinfo{person}{Richard Cyganiak}, {and}
  \bibinfo{person}{Zachary~G. Ives}.} \bibinfo{year}{2007}\natexlab{}.
\newblock \showarticletitle{DBpedia: {A} Nucleus for a Web of Open Data}. In
  \bibinfo{booktitle}{\emph{The Semantic Web, 6th International Semantic Web
  Conference, 2nd Asian Semantic Web Conference, {ISWC} 2007 + {ASWC} 2007,
  Busan, Korea, November 11-15, 2007}} \emph{(\bibinfo{series}{Lecture Notes in
  Computer Science}, Vol.~\bibinfo{volume}{4825})},
  \bibfield{editor}{\bibinfo{person}{Karl Aberer}, \bibinfo{person}{Key{-}Sun
  Choi}, \bibinfo{person}{Natasha~Fridman Noy}, \bibinfo{person}{Dean
  Allemang}, \bibinfo{person}{Kyung{-}Il Lee}, \bibinfo{person}{Lyndon J.~B.
  Nixon}, \bibinfo{person}{Jennifer Golbeck}, \bibinfo{person}{Peter Mika},
  \bibinfo{person}{Diana Maynard}, \bibinfo{person}{Riichiro Mizoguchi},
  \bibinfo{person}{Guus Schreiber}, {and} \bibinfo{person}{Philippe
  Cudr{\'{e}}{-}Mauroux}} (Eds.). \bibinfo{publisher}{Springer},
  \bibinfo{pages}{722--735}.
\newblock
\urldef\tempurl%
\url{https://doi.org/10.1007/978-3-540-76298-0\_52}
\showDOI{\tempurl}


\bibitem[\protect\citeauthoryear{Bordes, Usunier, Garc{\'{\i}}a{-}Dur{\'{a}}n,
  Weston, and Yakhnenko}{Bordes et~al\mbox{.}}{2013}]%
        {DBLP:conf/nips/BordesUGWY13}
\bibfield{author}{\bibinfo{person}{Antoine Bordes}, \bibinfo{person}{Nicolas
  Usunier}, \bibinfo{person}{Alberto Garc{\'{\i}}a{-}Dur{\'{a}}n},
  \bibinfo{person}{Jason Weston}, {and} \bibinfo{person}{Oksana Yakhnenko}.}
  \bibinfo{year}{2013}\natexlab{}.
\newblock \showarticletitle{Translating Embeddings for Modeling
  Multi-relational Data}. In \bibinfo{booktitle}{\emph{Advances in Neural
  Information Processing Systems 26: 27th Annual Conference on Neural
  Information Processing Systems 2013. Proceedings of a meeting held December
  5-8, 2013, Lake Tahoe, Nevada, United States}},
  \bibfield{editor}{\bibinfo{person}{Christopher J.~C. Burges},
  \bibinfo{person}{L{\'{e}}on Bottou}, \bibinfo{person}{Zoubin Ghahramani},
  {and} \bibinfo{person}{Kilian~Q. Weinberger}} (Eds.).
  \bibinfo{pages}{2787--2795}.
\newblock
\urldef\tempurl%
\url{http://papers.nips.cc/paper/5071-translating-embeddings-for-modeling-multi-relational-data}
\showURL{%
\tempurl}


\bibitem[\protect\citeauthoryear{Cao, Liu, Li, Liu, Li, and Chua}{Cao
  et~al\mbox{.}}{2019}]%
        {DBLP:conf/acl/CaoLLLLC19}
\bibfield{author}{\bibinfo{person}{Yixin Cao}, \bibinfo{person}{Zhiyuan Liu},
  \bibinfo{person}{Chengjiang Li}, \bibinfo{person}{Zhiyuan Liu},
  \bibinfo{person}{Juanzi Li}, {and} \bibinfo{person}{Tat{-}Seng Chua}.}
  \bibinfo{year}{2019}\natexlab{}.
\newblock \showarticletitle{Multi-Channel Graph Neural Network for Entity
  Alignment}. In \bibinfo{booktitle}{\emph{Proceedings of the 57th Conference
  of the Association for Computational Linguistics, {ACL} 2019, Florence,
  Italy, July 28- August 2, 2019, Volume 1: Long Papers}}.
  \bibinfo{pages}{1452--1461}.
\newblock
\urldef\tempurl%
\url{https://doi.org/10.18653/v1/p19-1140}
\showDOI{\tempurl}


\bibitem[\protect\citeauthoryear{Chen, Tian, Yang, and Zaniolo}{Chen
  et~al\mbox{.}}{2017}]%
        {DBLP:conf/ijcai/ChenTYZ17}
\bibfield{author}{\bibinfo{person}{Muhao Chen}, \bibinfo{person}{Yingtao Tian},
  \bibinfo{person}{Mohan Yang}, {and} \bibinfo{person}{Carlo Zaniolo}.}
  \bibinfo{year}{2017}\natexlab{}.
\newblock \showarticletitle{Multilingual Knowledge Graph Embeddings for
  Cross-lingual Knowledge Alignment}. In \bibinfo{booktitle}{\emph{Proceedings
  of the Twenty-Sixth International Joint Conference on Artificial
  Intelligence, {IJCAI} 2017, Melbourne, Australia, August 19-25, 2017}}.
  \bibinfo{pages}{1511--1517}.
\newblock
\urldef\tempurl%
\url{https://doi.org/10.24963/ijcai.2017/209}
\showDOI{\tempurl}


\bibitem[\protect\citeauthoryear{Guo, Sun, and Hu}{Guo et~al\mbox{.}}{2019}]%
        {DBLP:conf/icml/GuoSH19}
\bibfield{author}{\bibinfo{person}{Lingbing Guo}, \bibinfo{person}{Zequn Sun},
  {and} \bibinfo{person}{Wei Hu}.} \bibinfo{year}{2019}\natexlab{}.
\newblock \showarticletitle{Learning to Exploit Long-term Relational
  Dependencies in Knowledge Graphs}. In \bibinfo{booktitle}{\emph{Proceedings
  of the 36th International Conference on Machine Learning, {ICML} 2019, 9-15
  June 2019, Long Beach, California, {USA}}}. \bibinfo{pages}{2505--2514}.
\newblock
\urldef\tempurl%
\url{http://proceedings.mlr.press/v97/guo19c.html}
\showURL{%
\tempurl}


\bibitem[\protect\citeauthoryear{Hadsell, Chopra, and LeCun}{Hadsell
  et~al\mbox{.}}{2006}]%
        {DBLP:conf/cvpr/HadsellCL06}
\bibfield{author}{\bibinfo{person}{Raia Hadsell}, \bibinfo{person}{Sumit
  Chopra}, {and} \bibinfo{person}{Yann LeCun}.}
  \bibinfo{year}{2006}\natexlab{}.
\newblock \showarticletitle{Dimensionality Reduction by Learning an Invariant
  Mapping}. In \bibinfo{booktitle}{\emph{2006 {IEEE} Computer Society
  Conference on Computer Vision and Pattern Recognition {(CVPR} 2006), 17-22
  June 2006, New York, NY, {USA}}}. \bibinfo{pages}{1735--1742}.
\newblock
\urldef\tempurl%
\url{https://doi.org/10.1109/CVPR.2006.100}
\showDOI{\tempurl}


\bibitem[\protect\citeauthoryear{He, Zhang, Ren, and Sun}{He
  et~al\mbox{.}}{2015}]%
        {DBLP:conf/iccv/HeZRS15}
\bibfield{author}{\bibinfo{person}{Kaiming He}, \bibinfo{person}{Xiangyu
  Zhang}, \bibinfo{person}{Shaoqing Ren}, {and} \bibinfo{person}{Jian Sun}.}
  \bibinfo{year}{2015}\natexlab{}.
\newblock \showarticletitle{Delving Deep into Rectifiers: Surpassing
  Human-Level Performance on ImageNet Classification}. In
  \bibinfo{booktitle}{\emph{2015 {IEEE} International Conference on Computer
  Vision, {ICCV} 2015, Santiago, Chile, December 7-13, 2015}}.
  \bibinfo{publisher}{{IEEE} Computer Society}, \bibinfo{pages}{1026--1034}.
\newblock
\urldef\tempurl%
\url{https://doi.org/10.1109/ICCV.2015.123}
\showDOI{\tempurl}


\bibitem[\protect\citeauthoryear{Ioffe and Szegedy}{Ioffe and Szegedy}{2015}]%
        {DBLP:conf/icml/IoffeS15}
\bibfield{author}{\bibinfo{person}{Sergey Ioffe} {and}
  \bibinfo{person}{Christian Szegedy}.} \bibinfo{year}{2015}\natexlab{}.
\newblock \showarticletitle{Batch Normalization: Accelerating Deep Network
  Training by Reducing Internal Covariate Shift}. In
  \bibinfo{booktitle}{\emph{Proceedings of the 32nd International Conference on
  Machine Learning, {ICML} 2015, Lille, France, 6-11 July 2015}}.
  \bibinfo{pages}{448--456}.
\newblock
\urldef\tempurl%
\url{http://proceedings.mlr.press/v37/ioffe15.html}
\showURL{%
\tempurl}


\bibitem[\protect\citeauthoryear{Kipf and Welling}{Kipf and Welling}{2016}]%
        {DBLP:journals/corr/KipfW16}
\bibfield{author}{\bibinfo{person}{Thomas~N. Kipf} {and} \bibinfo{person}{Max
  Welling}.} \bibinfo{year}{2016}\natexlab{}.
\newblock \showarticletitle{Semi-Supervised Classification with Graph
  Convolutional Networks}.
\newblock \bibinfo{journal}{\emph{CoRR}}  \bibinfo{volume}{abs/1609.02907}
  (\bibinfo{year}{2016}).
\newblock
\showeprint[arxiv]{1609.02907}
\urldef\tempurl%
\url{http://arxiv.org/abs/1609.02907}
\showURL{%
\tempurl}


\bibitem[\protect\citeauthoryear{Lample, Conneau, Ranzato, Denoyer, and
  J{\'{e}}gou}{Lample et~al\mbox{.}}{2018}]%
        {DBLP:conf/iclr/LampleCRDJ18}
\bibfield{author}{\bibinfo{person}{Guillaume Lample}, \bibinfo{person}{Alexis
  Conneau}, \bibinfo{person}{Marc'Aurelio Ranzato}, \bibinfo{person}{Ludovic
  Denoyer}, {and} \bibinfo{person}{Herv{\'{e}} J{\'{e}}gou}.}
  \bibinfo{year}{2018}\natexlab{}.
\newblock \showarticletitle{Word translation without parallel data}. In
  \bibinfo{booktitle}{\emph{6th International Conference on Learning
  Representations, {ICLR} 2018, Vancouver, BC, Canada, April 30 - May 3, 2018,
  Conference Track Proceedings}}. \bibinfo{publisher}{OpenReview.net}.
\newblock
\urldef\tempurl%
\url{https://openreview.net/forum?id=H196sainb}
\showURL{%
\tempurl}


\bibitem[\protect\citeauthoryear{Li, Cao, Hou, Shi, Li, and Chua}{Li
  et~al\mbox{.}}{2019a}]%
        {DBLP:conf/emnlp/LiCHSLC19}
\bibfield{author}{\bibinfo{person}{Chengjiang Li}, \bibinfo{person}{Yixin Cao},
  \bibinfo{person}{Lei Hou}, \bibinfo{person}{Jiaxin Shi},
  \bibinfo{person}{Juanzi Li}, {and} \bibinfo{person}{Tat{-}Seng Chua}.}
  \bibinfo{year}{2019}\natexlab{a}.
\newblock \showarticletitle{Semi-supervised Entity Alignment via Joint
  Knowledge Embedding Model and Cross-graph Model}. In
  \bibinfo{booktitle}{\emph{Proceedings of the 2019 Conference on Empirical
  Methods in Natural Language Processing and the 9th International Joint
  Conference on Natural Language Processing, {EMNLP-IJCNLP} 2019, Hong Kong,
  China, November 3-7, 2019}}. \bibinfo{pages}{2723--2732}.
\newblock
\urldef\tempurl%
\url{https://doi.org/10.18653/v1/D19-1274}
\showDOI{\tempurl}


\bibitem[\protect\citeauthoryear{Li, Gu, Dullien, Vinyals, and Kohli}{Li
  et~al\mbox{.}}{2019b}]%
        {DBLP:conf/icml/LiGDVK19}
\bibfield{author}{\bibinfo{person}{Yujia Li}, \bibinfo{person}{Chenjie Gu},
  \bibinfo{person}{Thomas Dullien}, \bibinfo{person}{Oriol Vinyals}, {and}
  \bibinfo{person}{Pushmeet Kohli}.} \bibinfo{year}{2019}\natexlab{b}.
\newblock \showarticletitle{Graph Matching Networks for Learning the Similarity
  of Graph Structured Objects}. In \bibinfo{booktitle}{\emph{Proceedings of the
  36th International Conference on Machine Learning, {ICML} 2019, 9-15 June
  2019, Long Beach, California, {USA}}}. \bibinfo{pages}{3835--3845}.
\newblock
\urldef\tempurl%
\url{http://proceedings.mlr.press/v97/li19d.html}
\showURL{%
\tempurl}


\bibitem[\protect\citeauthoryear{Lin, Liu, Sun, Liu, and Zhu}{Lin
  et~al\mbox{.}}{2015}]%
        {DBLP:conf/aaai/LinLSLZ15}
\bibfield{author}{\bibinfo{person}{Yankai Lin}, \bibinfo{person}{Zhiyuan Liu},
  \bibinfo{person}{Maosong Sun}, \bibinfo{person}{Yang Liu}, {and}
  \bibinfo{person}{Xuan Zhu}.} \bibinfo{year}{2015}\natexlab{}.
\newblock \showarticletitle{Learning Entity and Relation Embeddings for
  Knowledge Graph Completion}. In \bibinfo{booktitle}{\emph{Proceedings of the
  Twenty-Ninth {AAAI} Conference on Artificial Intelligence, January 25-30,
  2015, Austin, Texas, {USA}}}. \bibinfo{pages}{2181--2187}.
\newblock
\urldef\tempurl%
\url{http://www.aaai.org/ocs/index.php/AAAI/AAAI15/paper/view/9571}
\showURL{%
\tempurl}


\bibitem[\protect\citeauthoryear{Mao, Wang, Xu, Lan, and Wu}{Mao
  et~al\mbox{.}}{2020a}]%
        {DBLP:conf/wsdm/MaoWXLW20}
\bibfield{author}{\bibinfo{person}{Xin Mao}, \bibinfo{person}{Wenting Wang},
  \bibinfo{person}{Huimin Xu}, \bibinfo{person}{Man Lan}, {and}
  \bibinfo{person}{Yuanbin Wu}.} \bibinfo{year}{2020}\natexlab{a}.
\newblock \showarticletitle{{MRAEA:} An Efficient and Robust Entity Alignment
  Approach for Cross-lingual Knowledge Graph}. In
  \bibinfo{booktitle}{\emph{{WSDM} '20: The Thirteenth {ACM} International
  Conference on Web Search and Data Mining, Houston, TX, USA, February 3-7,
  2020}}. \bibinfo{pages}{420--428}.
\newblock
\urldef\tempurl%
\url{https://doi.org/10.1145/3336191.3371804}
\showDOI{\tempurl}


\bibitem[\protect\citeauthoryear{Mao, Wang, Xu, Wu, and Lan}{Mao
  et~al\mbox{.}}{2020b}]%
        {DBLP:conf/cikm/MaoWXWL20}
\bibfield{author}{\bibinfo{person}{Xin Mao}, \bibinfo{person}{Wenting Wang},
  \bibinfo{person}{Huimin Xu}, \bibinfo{person}{Yuanbin Wu}, {and}
  \bibinfo{person}{Man Lan}.} \bibinfo{year}{2020}\natexlab{b}.
\newblock \showarticletitle{Relational Reflection Entity Alignment}. In
  \bibinfo{booktitle}{\emph{{CIKM} '20: The 29th {ACM} International Conference
  on Information and Knowledge Management, Virtual Event, Ireland, October
  19-23, 2020}}. \bibinfo{pages}{1095--1104}.
\newblock
\urldef\tempurl%
\url{https://doi.org/10.1145/3340531.3412001}
\showDOI{\tempurl}


\bibitem[\protect\citeauthoryear{Schroff, Kalenichenko, and Philbin}{Schroff
  et~al\mbox{.}}{2015}]%
        {DBLP:conf/cvpr/SchroffKP15}
\bibfield{author}{\bibinfo{person}{Florian Schroff}, \bibinfo{person}{Dmitry
  Kalenichenko}, {and} \bibinfo{person}{James Philbin}.}
  \bibinfo{year}{2015}\natexlab{}.
\newblock \showarticletitle{FaceNet: {A} unified embedding for face recognition
  and clustering}. In \bibinfo{booktitle}{\emph{{IEEE} Conference on Computer
  Vision and Pattern Recognition, {CVPR} 2015, Boston, MA, USA, June 7-12,
  2015}}. \bibinfo{publisher}{{IEEE} Computer Society},
  \bibinfo{pages}{815--823}.
\newblock
\urldef\tempurl%
\url{https://doi.org/10.1109/CVPR.2015.7298682}
\showDOI{\tempurl}


\bibitem[\protect\citeauthoryear{Song, Xiang, Jegelka, and Savarese}{Song
  et~al\mbox{.}}{2016}]%
        {DBLP:conf/cvpr/SongXJS16}
\bibfield{author}{\bibinfo{person}{Hyun~Oh Song}, \bibinfo{person}{Yu Xiang},
  \bibinfo{person}{Stefanie Jegelka}, {and} \bibinfo{person}{Silvio Savarese}.}
  \bibinfo{year}{2016}\natexlab{}.
\newblock \showarticletitle{Deep Metric Learning via Lifted Structured Feature
  Embedding}. In \bibinfo{booktitle}{\emph{2016 {IEEE} Conference on Computer
  Vision and Pattern Recognition, {CVPR} 2016, Las Vegas, NV, USA, June 27-30,
  2016}}. \bibinfo{pages}{4004--4012}.
\newblock
\urldef\tempurl%
\url{https://doi.org/10.1109/CVPR.2016.434}
\showDOI{\tempurl}


\bibitem[\protect\citeauthoryear{Srivastava, Greff, and Schmidhuber}{Srivastava
  et~al\mbox{.}}{2015}]%
        {DBLP:journals/corr/SrivastavaGS15}
\bibfield{author}{\bibinfo{person}{Rupesh~Kumar Srivastava},
  \bibinfo{person}{Klaus Greff}, {and} \bibinfo{person}{J{\"{u}}rgen
  Schmidhuber}.} \bibinfo{year}{2015}\natexlab{}.
\newblock \showarticletitle{Highway Networks}.
\newblock \bibinfo{journal}{\emph{CoRR}}  \bibinfo{volume}{abs/1505.00387}
  (\bibinfo{year}{2015}).
\newblock
\showeprint[arxiv]{1505.00387}
\urldef\tempurl%
\url{http://arxiv.org/abs/1505.00387}
\showURL{%
\tempurl}


\bibitem[\protect\citeauthoryear{Suchanek, Kasneci, and Weikum}{Suchanek
  et~al\mbox{.}}{2007}]%
        {DBLP:conf/www/SuchanekKW07}
\bibfield{author}{\bibinfo{person}{Fabian~M. Suchanek},
  \bibinfo{person}{Gjergji Kasneci}, {and} \bibinfo{person}{Gerhard Weikum}.}
  \bibinfo{year}{2007}\natexlab{}.
\newblock \showarticletitle{Yago: a core of semantic knowledge}. In
  \bibinfo{booktitle}{\emph{Proceedings of the 16th International Conference on
  World Wide Web, {WWW} 2007, Banff, Alberta, Canada, May 8-12, 2007}},
  \bibfield{editor}{\bibinfo{person}{Carey~L. Williamson},
  \bibinfo{person}{Mary~Ellen Zurko}, \bibinfo{person}{Peter~F.
  Patel{-}Schneider}, {and} \bibinfo{person}{Prashant~J. Shenoy}} (Eds.).
  \bibinfo{publisher}{{ACM}}, \bibinfo{pages}{697--706}.
\newblock
\urldef\tempurl%
\url{https://doi.org/10.1145/1242572.1242667}
\showDOI{\tempurl}


\bibitem[\protect\citeauthoryear{Sun, Cheng, Zhang, Zhang, Zheng, Wang, and
  Wei}{Sun et~al\mbox{.}}{2020a}]%
        {DBLP:conf/cvpr/SunCZZZWW20}
\bibfield{author}{\bibinfo{person}{Yifan Sun}, \bibinfo{person}{Changmao
  Cheng}, \bibinfo{person}{Yuhan Zhang}, \bibinfo{person}{Chi Zhang},
  \bibinfo{person}{Liang Zheng}, \bibinfo{person}{Zhongdao Wang}, {and}
  \bibinfo{person}{Yichen Wei}.} \bibinfo{year}{2020}\natexlab{a}.
\newblock \showarticletitle{Circle Loss: {A} Unified Perspective of Pair
  Similarity Optimization}. In \bibinfo{booktitle}{\emph{2020 {IEEE/CVF}
  Conference on Computer Vision and Pattern Recognition, {CVPR} 2020, Seattle,
  WA, USA, June 13-19, 2020}}. \bibinfo{pages}{6397--6406}.
\newblock
\urldef\tempurl%
\url{https://doi.org/10.1109/CVPR42600.2020.00643}
\showDOI{\tempurl}


\bibitem[\protect\citeauthoryear{Sun, Hu, and Li}{Sun et~al\mbox{.}}{2017}]%
        {DBLP:conf/semweb/SunHL17}
\bibfield{author}{\bibinfo{person}{Zequn Sun}, \bibinfo{person}{Wei Hu}, {and}
  \bibinfo{person}{Chengkai Li}.} \bibinfo{year}{2017}\natexlab{}.
\newblock \showarticletitle{Cross-Lingual Entity Alignment via Joint
  Attribute-Preserving Embedding}. In \bibinfo{booktitle}{\emph{The Semantic
  Web - {ISWC} 2017 - 16th International Semantic Web Conference, Vienna,
  Austria, October 21-25, 2017, Proceedings, Part {I}}}
  \emph{(\bibinfo{series}{Lecture Notes in Computer Science},
  Vol.~\bibinfo{volume}{10587})}, \bibfield{editor}{\bibinfo{person}{Claudia
  d'Amato}, \bibinfo{person}{Miriam Fern{\'{a}}ndez},
  \bibinfo{person}{Valentina A.~M. Tamma}, \bibinfo{person}{Freddy
  L{\'{e}}cu{\'{e}}}, \bibinfo{person}{Philippe Cudr{\'{e}}{-}Mauroux},
  \bibinfo{person}{Juan~F. Sequeda}, \bibinfo{person}{Christoph Lange}, {and}
  \bibinfo{person}{Jeff Heflin}} (Eds.). \bibinfo{publisher}{Springer},
  \bibinfo{pages}{628--644}.
\newblock
\urldef\tempurl%
\url{https://doi.org/10.1007/978-3-319-68288-4\_37}
\showDOI{\tempurl}


\bibitem[\protect\citeauthoryear{Sun, Hu, Zhang, and Qu}{Sun
  et~al\mbox{.}}{2018}]%
        {DBLP:conf/ijcai/SunHZQ18}
\bibfield{author}{\bibinfo{person}{Zequn Sun}, \bibinfo{person}{Wei Hu},
  \bibinfo{person}{Qingheng Zhang}, {and} \bibinfo{person}{Yuzhong Qu}.}
  \bibinfo{year}{2018}\natexlab{}.
\newblock \showarticletitle{Bootstrapping Entity Alignment with Knowledge Graph
  Embedding}. In \bibinfo{booktitle}{\emph{Proceedings of the Twenty-Seventh
  International Joint Conference on Artificial Intelligence, {IJCAI} 2018, July
  13-19, 2018, Stockholm, Sweden}}. \bibinfo{pages}{4396--4402}.
\newblock
\urldef\tempurl%
\url{https://doi.org/10.24963/ijcai.2018/611}
\showDOI{\tempurl}


\bibitem[\protect\citeauthoryear{Sun, Huang, Hu, Chen, Guo, and Qu}{Sun
  et~al\mbox{.}}{2020b}]%
        {DBLP:journals/corr/abs-2004-13579}
\bibfield{author}{\bibinfo{person}{Zequn Sun}, \bibinfo{person}{JiaCheng
  Huang}, \bibinfo{person}{Wei Hu}, \bibinfo{person}{Muchao Chen},
  \bibinfo{person}{Lingbing Guo}, {and} \bibinfo{person}{Yuzhong Qu}.}
  \bibinfo{year}{2020}\natexlab{b}.
\newblock \showarticletitle{TransEdge: Translating Relation-contextualized
  Embeddings for Knowledge Graphs}.
\newblock \bibinfo{journal}{\emph{CoRR}}  \bibinfo{volume}{abs/2004.13579}
  (\bibinfo{year}{2020}).
\newblock
\showeprint[arxiv]{2004.13579}
\urldef\tempurl%
\url{https://arxiv.org/abs/2004.13579}
\showURL{%
\tempurl}


\bibitem[\protect\citeauthoryear{Sun, Wang, Hu, Chen, Dai, Zhang, and Qu}{Sun
  et~al\mbox{.}}{2020c}]%
        {DBLP:conf/aaai/SunW0CDZQ20}
\bibfield{author}{\bibinfo{person}{Zequn Sun}, \bibinfo{person}{Chengming
  Wang}, \bibinfo{person}{Wei Hu}, \bibinfo{person}{Muhao Chen},
  \bibinfo{person}{Jian Dai}, \bibinfo{person}{Wei Zhang}, {and}
  \bibinfo{person}{Yuzhong Qu}.} \bibinfo{year}{2020}\natexlab{c}.
\newblock \showarticletitle{Knowledge Graph Alignment Network with Gated
  Multi-Hop Neighborhood Aggregation}. In \bibinfo{booktitle}{\emph{The
  Thirty-Fourth {AAAI} Conference on Artificial Intelligence, {AAAI} 2020, The
  Thirty-Second Innovative Applications of Artificial Intelligence Conference,
  {IAAI} 2020, The Tenth {AAAI} Symposium on Educational Advances in Artificial
  Intelligence, {EAAI} 2020, New York, NY, USA, February 7-12, 2020}}.
  \bibinfo{pages}{222--229}.
\newblock
\urldef\tempurl%
\url{https://aaai.org/ojs/index.php/AAAI/article/view/5354}
\showURL{%
\tempurl}


\bibitem[\protect\citeauthoryear{Tang, Zhang, Yao, Li, Zhang, and Su}{Tang
  et~al\mbox{.}}{2008}]%
        {DBLP:conf/kdd/TangZYLZS08}
\bibfield{author}{\bibinfo{person}{Jie Tang}, \bibinfo{person}{Jing Zhang},
  \bibinfo{person}{Limin Yao}, \bibinfo{person}{Juanzi Li}, \bibinfo{person}{Li
  Zhang}, {and} \bibinfo{person}{Zhong Su}.} \bibinfo{year}{2008}\natexlab{}.
\newblock \showarticletitle{ArnetMiner: extraction and mining of academic
  social networks}. In \bibinfo{booktitle}{\emph{Proceedings of the 14th {ACM}
  {SIGKDD} International Conference on Knowledge Discovery and Data Mining, Las
  Vegas, Nevada, USA, August 24-27, 2008}},
  \bibfield{editor}{\bibinfo{person}{Ying Li}, \bibinfo{person}{Bing Liu},
  {and} \bibinfo{person}{Sunita Sarawagi}} (Eds.). \bibinfo{publisher}{{ACM}},
  \bibinfo{pages}{990--998}.
\newblock
\urldef\tempurl%
\url{https://doi.org/10.1145/1401890.1402008}
\showDOI{\tempurl}


\bibitem[\protect\citeauthoryear{Velickovic, Cucurull, Casanova, Romero,
  Li{\`{o}}, and Bengio}{Velickovic et~al\mbox{.}}{2018}]%
        {DBLP:conf/iclr/VelickovicCCRLB18}
\bibfield{author}{\bibinfo{person}{Petar Velickovic}, \bibinfo{person}{Guillem
  Cucurull}, \bibinfo{person}{Arantxa Casanova}, \bibinfo{person}{Adriana
  Romero}, \bibinfo{person}{Pietro Li{\`{o}}}, {and} \bibinfo{person}{Yoshua
  Bengio}.} \bibinfo{year}{2018}\natexlab{}.
\newblock \showarticletitle{Graph Attention Networks}. In
  \bibinfo{booktitle}{\emph{6th International Conference on Learning
  Representations, {ICLR} 2018, Vancouver, BC, Canada, April 30 - May 3, 2018,
  Conference Track Proceedings}}.
\newblock
\urldef\tempurl%
\url{https://openreview.net/forum?id=rJXMpikCZ}
\showURL{%
\tempurl}


\bibitem[\protect\citeauthoryear{Wang, Lv, Lan, and Zhang}{Wang
  et~al\mbox{.}}{2018}]%
        {DBLP:conf/emnlp/WangLLZ18}
\bibfield{author}{\bibinfo{person}{Zhichun Wang}, \bibinfo{person}{Qingsong
  Lv}, \bibinfo{person}{Xiaohan Lan}, {and} \bibinfo{person}{Yu Zhang}.}
  \bibinfo{year}{2018}\natexlab{}.
\newblock \showarticletitle{Cross-lingual Knowledge Graph Alignment via Graph
  Convolutional Networks}. In \bibinfo{booktitle}{\emph{Proceedings of the 2018
  Conference on Empirical Methods in Natural Language Processing, Brussels,
  Belgium, October 31 - November 4, 2018}}. \bibinfo{pages}{349--357}.
\newblock
\urldef\tempurl%
\url{https://doi.org/10.18653/v1/d18-1032}
\showDOI{\tempurl}


\bibitem[\protect\citeauthoryear{Wang, Zhang, Feng, and Chen}{Wang
  et~al\mbox{.}}{2014}]%
        {DBLP:conf/aaai/WangZFC14}
\bibfield{author}{\bibinfo{person}{Zhen Wang}, \bibinfo{person}{Jianwen Zhang},
  \bibinfo{person}{Jianlin Feng}, {and} \bibinfo{person}{Zheng Chen}.}
  \bibinfo{year}{2014}\natexlab{}.
\newblock \showarticletitle{Knowledge Graph Embedding by Translating on
  Hyperplanes}. In \bibinfo{booktitle}{\emph{Proceedings of the Twenty-Eighth
  {AAAI} Conference on Artificial Intelligence, July 27 -31, 2014, Qu{\'{e}}bec
  City, Qu{\'{e}}bec, Canada}}. \bibinfo{pages}{1112--1119}.
\newblock
\urldef\tempurl%
\url{http://www.aaai.org/ocs/index.php/AAAI/AAAI14/paper/view/8531}
\showURL{%
\tempurl}


\bibitem[\protect\citeauthoryear{Wu, Liu, Feng, Wang, Yan, and Zhao}{Wu
  et~al\mbox{.}}{2019b}]%
        {DBLP:conf/ijcai/WuLF0Y019}
\bibfield{author}{\bibinfo{person}{Yuting Wu}, \bibinfo{person}{Xiao Liu},
  \bibinfo{person}{Yansong Feng}, \bibinfo{person}{Zheng Wang},
  \bibinfo{person}{Rui Yan}, {and} \bibinfo{person}{Dongyan Zhao}.}
  \bibinfo{year}{2019}\natexlab{b}.
\newblock \showarticletitle{Relation-Aware Entity Alignment for Heterogeneous
  Knowledge Graphs}. In \bibinfo{booktitle}{\emph{Proceedings of the
  Twenty-Eighth International Joint Conference on Artificial Intelligence,
  {IJCAI} 2019, Macao, China, August 10-16, 2019}}.
  \bibinfo{pages}{5278--5284}.
\newblock
\urldef\tempurl%
\url{https://doi.org/10.24963/ijcai.2019/733}
\showDOI{\tempurl}


\bibitem[\protect\citeauthoryear{Wu, Liu, Feng, Wang, and Zhao}{Wu
  et~al\mbox{.}}{2019a}]%
        {DBLP:conf/emnlp/WuLFWZ19}
\bibfield{author}{\bibinfo{person}{Yuting Wu}, \bibinfo{person}{Xiao Liu},
  \bibinfo{person}{Yansong Feng}, \bibinfo{person}{Zheng Wang}, {and}
  \bibinfo{person}{Dongyan Zhao}.} \bibinfo{year}{2019}\natexlab{a}.
\newblock \showarticletitle{Jointly Learning Entity and Relation
  Representations for Entity Alignment}. In
  \bibinfo{booktitle}{\emph{Proceedings of the 2019 Conference on Empirical
  Methods in Natural Language Processing and the 9th International Joint
  Conference on Natural Language Processing, {EMNLP-IJCNLP} 2019, Hong Kong,
  China, November 3-7, 2019}}. \bibinfo{pages}{240--249}.
\newblock
\urldef\tempurl%
\url{https://doi.org/10.18653/v1/D19-1023}
\showDOI{\tempurl}


\bibitem[\protect\citeauthoryear{Xu, Wang, Yu, Feng, Song, Wang, and Yu}{Xu
  et~al\mbox{.}}{2019}]%
        {DBLP:conf/acl/XuWYFSWY19}
\bibfield{author}{\bibinfo{person}{Kun Xu}, \bibinfo{person}{Liwei Wang},
  \bibinfo{person}{Mo Yu}, \bibinfo{person}{Yansong Feng}, \bibinfo{person}{Yan
  Song}, \bibinfo{person}{Zhiguo Wang}, {and} \bibinfo{person}{Dong Yu}.}
  \bibinfo{year}{2019}\natexlab{}.
\newblock \showarticletitle{Cross-lingual Knowledge Graph Alignment via Graph
  Matching Neural Network}. In \bibinfo{booktitle}{\emph{Proceedings of the
  57th Conference of the Association for Computational Linguistics, {ACL} 2019,
  Florence, Italy, July 28- August 2, 2019, Volume 1: Long Papers}}.
  \bibinfo{pages}{3156--3161}.
\newblock
\urldef\tempurl%
\url{https://doi.org/10.18653/v1/p19-1304}
\showDOI{\tempurl}


\bibitem[\protect\citeauthoryear{Yang, Zou, Shi, Lu, Lin, and Sun}{Yang
  et~al\mbox{.}}{2019}]%
        {DBLP:conf/emnlp/YangZSLLS19}
\bibfield{author}{\bibinfo{person}{Hsiu{-}Wei Yang}, \bibinfo{person}{Yanyan
  Zou}, \bibinfo{person}{Peng Shi}, \bibinfo{person}{Wei Lu},
  \bibinfo{person}{Jimmy Lin}, {and} \bibinfo{person}{Xu Sun}.}
  \bibinfo{year}{2019}\natexlab{}.
\newblock \showarticletitle{Aligning Cross-Lingual Entities with Multi-Aspect
  Information}. In \bibinfo{booktitle}{\emph{Proceedings of the 2019 Conference
  on Empirical Methods in Natural Language Processing and the 9th International
  Joint Conference on Natural Language Processing, {EMNLP-IJCNLP} 2019, Hong
  Kong, China, November 3-7, 2019}}. \bibinfo{pages}{4430--4440}.
\newblock
\urldef\tempurl%
\url{https://doi.org/10.18653/v1/D19-1451}
\showDOI{\tempurl}


\bibitem[\protect\citeauthoryear{Yun, Jeong, Kim, Kang, and Kim}{Yun
  et~al\mbox{.}}{2019}]%
        {DBLP:conf/nips/YunJKKK19}
\bibfield{author}{\bibinfo{person}{Seongjun Yun}, \bibinfo{person}{Minbyul
  Jeong}, \bibinfo{person}{Raehyun Kim}, \bibinfo{person}{Jaewoo Kang}, {and}
  \bibinfo{person}{Hyunwoo~J. Kim}.} \bibinfo{year}{2019}\natexlab{}.
\newblock \showarticletitle{Graph Transformer Networks}. In
  \bibinfo{booktitle}{\emph{Advances in Neural Information Processing Systems
  32: Annual Conference on Neural Information Processing Systems 2019, NeurIPS
  2019, 8-14 December 2019, Vancouver, BC, Canada}}.
  \bibinfo{pages}{11960--11970}.
\newblock
\urldef\tempurl%
\url{http://papers.nips.cc/paper/9367-graph-transformer-networks}
\showURL{%
\tempurl}


\bibitem[\protect\citeauthoryear{{Zhao}, {Zeng}, {Tang}, {Wang}, and
  {Suchanek}}{{Zhao} et~al\mbox{.}}{2020}]%
        {9174835}
\bibfield{author}{\bibinfo{person}{X. {Zhao}}, \bibinfo{person}{W. {Zeng}},
  \bibinfo{person}{J. {Tang}}, \bibinfo{person}{W. {Wang}}, {and}
  \bibinfo{person}{F. {Suchanek}}.} \bibinfo{year}{2020}\natexlab{}.
\newblock \showarticletitle{An Experimental Study of State-of-the-Art Entity
  Alignment Approaches}.
\newblock \bibinfo{journal}{\emph{IEEE Transactions on Knowledge and Data
  Engineering}} (\bibinfo{year}{2020}), \bibinfo{pages}{1--1}.
\newblock


\bibitem[\protect\citeauthoryear{Zhu, Zhou, Wu, Tan, and Guo}{Zhu
  et~al\mbox{.}}{2019}]%
        {DBLP:conf/ijcai/ZhuZ0TG19}
\bibfield{author}{\bibinfo{person}{Qiannan Zhu}, \bibinfo{person}{Xiaofei
  Zhou}, \bibinfo{person}{Jia Wu}, \bibinfo{person}{Jianlong Tan}, {and}
  \bibinfo{person}{Li Guo}.} \bibinfo{year}{2019}\natexlab{}.
\newblock \showarticletitle{Neighborhood-Aware Attentional Representation for
  Multilingual Knowledge Graphs}. In \bibinfo{booktitle}{\emph{Proceedings of
  the Twenty-Eighth International Joint Conference on Artificial Intelligence,
  {IJCAI} 2019, Macao, China, August 10-16, 2019}}.
  \bibinfo{pages}{1943--1949}.
\newblock
\urldef\tempurl%
\url{https://doi.org/10.24963/ijcai.2019/269}
\showDOI{\tempurl}


\end{thebibliography}

\end{document}